\documentclass[graybox]{svmult}

\usepackage{mathptmx}       
\usepackage{helvet}         
\usepackage{courier}        
\usepackage{type1cm}        
\usepackage{makeidx}         
\usepackage{graphicx}        
\usepackage{multicol}        
\usepackage[bottom]{footmisc}

\makeindex             
\usepackage{amsmath,amsfonts,amssymb,latexsym,bbm}
\usepackage[greek,english]{babel}
\newcommand{\tg}[1]{\textgreek{#1}} 
\usepackage[utf8]{inputenc}
\usepackage{array}
\usepackage{tabularx}
\usepackage{url,subfigure}
\usepackage{epsfig}

\newcommand{\REAL}{\ensuremath{\mathbb{R}}}
\newcommand{\INT}{\ensuremath{\mathbb{Z}}}

\newcommand{\EREAL}{\ensuremath{\overline{\REAL}}}
\newcommand{\Rmax}{\ensuremath{\mathbb{R}_\textrm{max}}}
\newcommand{\Rmin}{\ensuremath{\mathbb{R}_\textrm{min}}}

\newcommand{\beq}{\begin{equation}}
\newcommand{\eeq}{\end{equation}}
\newcommand{\bea}{\begin{eqnarray}}
\newcommand{\eea}{\end{eqnarray}}
\newcommand{\bean}{\begin{eqnarray*}}
\newcommand{\eean}{\end{eqnarray*}}
\newcommand{\bcen}{\begin{center}}
\newcommand{\ecen}{\end{center}}
\newcommand{\bitm}{\begin{itemize}}
\newcommand{\eitm}{\end{itemize}}

\newtheorem{Theorem}{\bf Theorem}

\newtheorem{Example}{\bf Example}
\newcommand{\Hh}{\ensuremath{\mathcal{H}}}

\newcommand{\Kk}{\ensuremath{\mathcal{K}}}
\newcommand{\Ll}{\ensuremath{\mathcal{L}}}
\newcommand{\Mm}{\ensuremath{\mathcal{M}}}

\newcommand{\Ss}{\ensuremath{\mathcal{S}}}
\newcommand{\Tt}{\ensuremath{\mathcal{T}}}
\newcommand{\Vv}{\ensuremath{\mathcal{V}}}

\newcommand{\Ww}{\ensuremath{\mathcal{W}}}
\newcommand{\Xx}{\ensuremath{\mathcal{X}}}

\newcommand{\sbs}{\ensuremath{\subseteq \,}}    
\newcommand{\keno}{\ensuremath{\varnothing}}    
\newcommand{\defineq}{\ensuremath{\, \triangleq \,}}    
\newcommand{\defineql}{\ensuremath{\, := \,}}    

\newcommand{\dilt}{\ensuremath{\oplus}}
\newcommand{\eros}{\ensuremath{\ominus}}

\newcommand{\tran}[2]{\ensuremath{{#1}_{#2}}} 
\newcommand{\refl}[1]{\ensuremath{{#1}^s}} 

\newcommand{\flatdom}{\ensuremath{E}} 
\def\sgcnv{\hbox{$\, \bigcirc \,$\kern-0.9em\hbox{\mgop}$\,\,\,$}} 
\def\igcnv{\hbox{$\, \bigcirc \,$\kern-0.9em\hbox{\mgop}$\, '\,$}} 
\def\supgeno{\hbox{$\, \bigcirc \,$\kern-1.0em\hbox{$\wedge$}$\,$}} 
\def\infgeno{\hbox{$\, \bigcirc \,$\kern-1.0em\hbox{$\vee$}$\,$}} 

\newcommand{\pord}{\ensuremath{\leq}} 
\newcommand{\spord}{\ensuremath{<}} 
\newcommand{\ipord}{\ensuremath{\geq}} 
\newcommand{\dpord}{\ensuremath{\pord '}} 
\newcommand{\ltle}{\ensuremath{\bot}}  
\newcommand{\ltge}{\ensuremath{\top}}  
\newcommand{\flatle}{\ensuremath{O}}  
\newcommand{\flatge}{\ensuremath{I}}  
\newcommand{\dimpls}{\ensuremath{q}} 
\newcommand{\eimpls}{\ensuremath{q'}} 
\newcommand{\dlop}{\mbox{\large $\delta$}} 
\newcommand{\erop}{\mbox{\large $\varepsilon$}} 
\newcommand{\fdlop}{\ensuremath{\Delta}}  
\newcommand{\ferop}{\ensuremath{\mathcal{E}}}  
\newcommand{\sbdl}{\ensuremath{\eta}}  
\newcommand{\asbdl}{\ensuremath{\zeta}} 

\newcommand{\opop}{\mbox{\large $\alpha$}} 
\newcommand{\clop}{\mbox{\large $\beta$}} 
\newcommand{\idop}{\ensuremath{\mathbf{id}}} 
\newcommand{\trop}{\mbox{\large $\tau$}} 

\newcommand{\vset}{\ensuremath{\mathcal{K}}} 
\newcommand{\vsetle}{\ensuremath{\bot}} 
\newcommand{\vsetge}{\ensuremath{\top}} 
\newcommand{\mgid}{\ensuremath{e}} 
\newcommand{\dmgid}{\ensuremath{e'}} 
\newcommand{\vsgroup}{\ensuremath{G}} 
\newcommand{\mgop}{\ensuremath{\star}}  
\newcommand{\dmgop}{\ensuremath{\mgop'}} 

\newcommand{\vct}[1]{\ensuremath{\mathbf{#1}}}  
\newcommand{\mtr}[1]{\ensuremath{\mathbf{#1}}}  
\newcommand{\mxgmp}{\ensuremath{\ \frame{\mgop}\ }} 
\newcommand{\mngmp}{\ensuremath{\ \frame{\mgop}\, '\ }} 
\newcommand{\mxsmp}{\ensuremath{\boxplus}} 
\newcommand{\mnsmp}{\ensuremath{\mxsmp'}} 
\newcommand{\mnasbdmp}{\ensuremath{\Box'_{\asbdl}}}  

\newcommand{\glconj}[1]{\ensuremath{{#1}^\neg}} 
\newcommand{\eresid}[1]{\ensuremath{{#1}^\sharp}} 
\newcommand{\conjtranmtr}[1]{\ensuremath{{#1}^\ast}} 

\newcommand{\fun}{\ensuremath{\mathrm{Fun}}}    

\newcommand{\rank}{\ensuremath{\mathrm{rank}}}

\newcommand{\conv}{\ensuremath{\mathrm{conv}}} 
\newcommand{\newtpoly}{\ensuremath{\mathrm{New}}} 

\begin{document}
%
\title*{Tropical Geometry and Piecewise-Linear Approximation of Curves and Surfaces  on Weighted Lattices}
%
\titlerunning{Tropical Geometry,  Shape Approximation and Weighted Lattices}
%
\author{Petros Maragos and Emmanouil Theodosis}
\authorrunning{P. Maragos and E. Theodosis}
%
\institute{Petros Maragos \at National Technical University of Athens, School of
Electrical and Computer Engineering,  Athens 15773, Greece. \email{maragos@cs.ntua.gr}. \\
Emmanouil Theodosis \at Harvard University, School of Engineering and Applied Sciences, Cambridge, MA 02138, USA.
\email{etheodosis@g.harvard.edu}.
}
\maketitle              
\abstract{
 Tropical Geometry and Mathematical Morphology share the same max-plus and min-plus semiring arithmetic and  matrix algebra.
 In this chapter we summarize some of their main ideas and common (geometric and algebraic) structure, generalize and extend both of them using weighted lattices and a max-$\mgop$ algebra with an arbitrary binary operation $\mgop$ that distributes over max, and outline applications to geometry,  machine learning,   and optimization.  Further,
we  generalize tropical geometrical objects using weighted lattices.  Finally,
 we provide the optimal solution of max-$\mgop$ equations using morphological adjunctions that are projections on weighted lattices,  and apply it to optimal piecewise-linear regression for fitting max-$\mgop$ tropical curves and surfaces to arbitrary data that constitute polygonal or polyhedral shape approximations.
 This also includes an efficient algorithm for solving the convex regression problem of data fitting with max-affine functions.
}

\section{Introduction}

As stated in \cite{MaSt15}, tropical geometry is a ``marriage between algebraic geometry and polyhedral geometry". It is a relatively recent field in mathematics and computer science. However, the scalar arithmetic of its analytic part pre-existed in the form of max-plus and min-plus semiring arithmetic used in finite automata, nonlinear functional and image analysis, convex analysis, nonlinear control and optimization. In this chapter we explore the parts it shares with morphological image analysis, extend both of them using weighted lattices, and apply max-plus algebra to optimally fitting of tropical curves and surfaces to data.

%

Combinations of max-plus, or its dual min-plus, arithmetic with corresponding nonlinear matrix algebra and signal convolutions have been used in operations research and scheduling \cite{Cuni79}; discrete event systems, max-plus control and optimization
\cite{AGG12,AGNS11,BCOQ01,Butk10,CDQV85,Gaub97,GoMi08,HOW06,McEn06,BoSc12}; convex analysis \cite{Rock70,Luce10};
morphological image analysis \cite{Heij94,Mara05b,Meye19,Serr82,Serr88};
nonlinear PDEs of the Hamilton-Jacobi type and vision scale-spaces \cite{BrMa94,HeMa97};
speech recognition and natural language processing \cite{HoNa13,MPR02};
neural networks \cite{ChMa17,ChMa18,GWM+13,PeMa00,RiUr03,YaMa95,ZNL18,Zha+19};
idempotent mathematics (nonlinear functional analysis) \cite{LMS01,Litv07}.

Max-plus (a.k.a. `max-sum') arithmetic forms an idempotent semiring denoted as $(\Rmax, \max, +)$ where $\Rmax=\REAL \cup \{-\infty\}$ and the real number addition and multiplication are replaced by the max and sum operations, respectively. As an idempotent semiring it is covered by the theory of dioids \cite{GoMi08}. The dual min-plus semiring $(\Rmin, \min, +)$, where $\Rmin=\REAL\cup \{+\infty\}$, has been called `tropical semiring' and has been used in finite automata \cite{Peri98,Simo94}, speech and language recognition using graphical models \cite{MPR02},  and  tropical geometry \cite{MaSt15,Mikh05}.
In idempotent mathematics \cite{LMS01,Litv07}, in convex optimization \cite{BoVa04}, and in the general theory of dioids \cite{GoMi08}, they often use the limit of the \emph{Log-Sum-Exp} approximation for the max and min operations: 
\beq
\begin{array}{rcl}
\lim _{\theta \downarrow 0} \theta \cdot \log ({e^{a/\theta}} + {e^{b/\theta}}) & = & \max (a,b)\\
\lim _{\theta \downarrow 0} (-\theta)\log ({e^{ - a/\theta}} + {e^{ - b/\theta}}) & = & \min (a,b)
\end{array}
\label{lseapprox}
\eeq
where $\theta>0$ is usually called a `temperature' parameter. This approximation is at the heart of the `Maslov Dequantization' \cite{Masl87} of real numbers,
 which generates a whole family of semirings $S_\theta=(\Rmax,+_\theta, \times _\theta)$, $\theta>0$,
 whose operations of generalized  `addition' $+_\theta$ and `multiplication' $\times_\theta$ are defined as
\beq
\begin{array}{rcl}
a+_\theta b & \defineql & \theta \cdot \log ({e^{a/\theta}} + {e^{b/\theta}})
=\phi_\theta^{-1}[\phi _\theta(a)+\phi _\theta(b)]
\\
a\times _\theta b & \defineql & a+b
\end{array}
\label{lsesemiring}
\eeq
with $\phi _\theta(a)\defineql \exp(a/\theta)$.
This makes $S_\theta$ isomorphic to the semiring of nonnegative real numbers $\REAL_{\geq 0}$ equipped with standard addition and multiplication.
 This isomorphism is enabled via the logarithmic mapping $a\mapsto \phi_\theta^{-1}(a)=\theta \log (a)$.
 In the limit $\theta\downarrow 0$,  we get $S_0$ which is the max-plus semiring.
Thus, $(\REAL_{\geq 0},+,\times)$ is isomorphic to $(\Rmax,\max,+)$.

Max and min operations (or more generally supremum and infimum) form the algebra of lattices, which has been used  to generalize Euclidean morphology \cite{Serr82}, based on Minkowski set operations and their extensions to functions via level sets,  to more general morphological operators on complete lattices \cite{Serr88,Heij94,HeRo90,BaBa93}.
 The scalar arithmetic of morphology on functions has been mainly flat;
 a few exceptions include the max-plus convolutions  and related operations which  have appeared in morphological image analysis \cite{Heij94,Mara05a,Serr82,Ster86},
 image algebra \cite{RiWi01}, convex analysis and optimization \cite{BeKa61,Luce10},
and nonlinear dynamical systems \cite{BCOQ01,Mara17}.
 Such non-flat morphological operations and their generalizations to a max-$\mgop$ algebra have been systematized and extended using the theory of weighted lattices \cite{Mara13,Mara17}. This connects morphology with max-plus algebra and tropical geometry.

 Tropical Geometry and the standard image operators of Mathematical Morphology share the same max-plus and min-plus semiring arithmetic and  matrix algebra.
In this chapter, whose preliminary version is based on \cite{Mara19}, we begin in Section~\ref{sc-mmfl} with some elementary concepts from classic Euclidean image morphological operators based on Minkowski set and function addition, duality pairs in the form of adjunctions (a.k.a. residuation pairs)  and their formalization using lattice theory. Then in Section~\ref{sc-deqntpde} we show how approximation (\ref{lseapprox}) converts the linear heat PDE modeling the Gaussian scale-space in computer vision to PDEs generating multiscale max-plus morphological operators. We continue in Section~\ref{sc-elemtropgeo} with elementary concepts and objects of tropical geometry.
Then, in Section~\ref{sc-wl} we summarize the theory of weighted lattices, which form nonlinear vector spaces, and use them to extend the max-plus mathematical morphology and tropical geometry  using a max-$\mgop$ algebra with an arbitrary binary operation $\mgop$ that distributes over max.
Further, we  generalize tropical geometrical objects using weighted lattices.
Finally, in Section~\ref{sc-solmaxeqn}
 we outline the optimal solution of max-$\mgop$ equations using morphological adjunctions (a.k.a. residuation pairs) that are projections  on weighted lattices,  and apply it in Section~\ref{sc-tropregres} to optimal convex piecewise-linear regression for fitting max-$\mgop$ tropical curves and surfaces to arbitrary data that constitute polygonal or polyhedral shape approximations.
 Throughout the chapter, we also outline applications to numerical geometry,  machine learning,   and optimization.

\textbf{Notation}:
For maximum (or supremum) and minimum (or infimum) operations we use the well-established lattice-theoretic symbols of $\vee$ and $\wedge$.
 We do \emph{not} use the notation $(\oplus,\otimes )$ for $(\max,+)$ or $(\min,+)$ which is frequently used in max-plus algebra,
because in  image analysis i)~the symbol $\oplus$ is extensively used for Minkowski set operations and max-plus convolutions, and ii)~$\otimes$ is unnecessarily confusing compared to the classic symbol $+$ of addition.
We use roman letters for functions, signals and their arguments and greek letters mainly for operators.
Also, boldface roman letters for vectors (lowcase)  and matrices (capital).
If $\mtr{M}=[m_{ij}]$ is a matrix, its $(i,j)$-th element is  denoted as  $m_{ij}$ or
as  $[ \mtr{M} ] _{ij}$. Similarly,
$\vct{x}=[x_i]$ denotes a column vector, whose $i$-th element
is denoted as $[\vct{x}]_i$ or simply $x_i$.

\section{Elements of Max-plus Morphology and Flat Lattices}
\label{sc-mmfl}

We view images, signals, and vectors as elements of \emph{complete lattices} $(\Ll, \vee, \wedge)$, where $\Ll$ is the set of lattice elements equipped with two binary operations, $\vee$ and $\wedge$, which denote the lattice supremum and infimum respectively.
Each of these operations induces a partial ordering $\pord$; e.g. for any $X,Y\in \Ll$, $X\pord Y \Longleftrightarrow Y=X\vee Y$.
The lattice operations satisfy many properties, 
as summarized in Table~\ref{tb-wlat}.
Conversely, a set $\Ll$ equipped with two binary operations $\vee$ and $\wedge$
that satisfy  properties (L1,L1$'$)--(L5,L5$'$)
 is a lattice whose
supremum is $\vee$, the infimum is $\wedge$, and partial ordering $\pord$ is given by (L6).
Completeness means that the supremum and infimum of any (even infinite) subset of $\Ll$ exists and belongs to $\Ll$.
A lattice $(\Ll , \vee ,\wedge)$ contains two weaker substructures:
 a sup-semilattice $(\Ll ,\vee)$
that satisfies properties (L$1-$L4) 
and an inf-semilattice $(\Ll,\wedge)$ that satisfies properties (L1$'-$L4$'$).
Examples of complete lattices we use in computer vision include
 i)~the lattice of Euclidean shapes, i.e. subsets of $\REAL^n$, equipped with set union and intersection,
  and ii)~the lattice  $\fun(\flatdom, \EREAL)$ of  functions 
with (arbitrary) domain $\flatdom$ and values in $\EREAL=\REAL\cup \{-\infty,+\infty\}$,
equipped with the pointwise supremum and pointwise infimum of extended real numbers.
For data processing, we also consider operators  $\psi: \Ll \rightarrow \Mm$ between two complete lattices.
The set  of all such operators becomes a complete  lattice if equipped with
the supremum, infimum, and partial ordering  defined pointwise for the operators' outputs.
\\

%
%
%



\noindent
\textbf{Monotone  Operators}:\

%
%
%
%
A lattice operator $\psi: \Ll \rightarrow \Mm$ is called \textit{increasing} or \textit{isotone}
 if it is order preserving; i.e. if, for any $X,Y\in \Ll$,  $X\pord Y \Longrightarrow \psi(X)\pord \psi (Y)$. (We use the same symbol for the partial order in $\Ll$ and $\Mm$ although they may be different,
 hoping that the difference will be clear from the context.)
Examples of increasing operators are the lattice homomorphisms
which preserve suprema and infima. 
If a lattice homomorphism  is also a bijection,
then it becomes an automorphism.
%
Four fundamental types of  increasing operators  are:
\emph{dilations} $\dlop$ and \emph{erosions} $\erop$
that satisfy respectively $\dlop (\bigvee _{i} X_i ) =\bigvee _{i} \dlop (X_i)$
and $\erop (\bigwedge _{i} X_i ) =\bigwedge _{i} \erop (X_i)$
over arbitrary (possibly infinite) collections;
 \emph{openings} $\opop$ that are increasing, idempotent ($\opop^2=\opop$), and antiextensive $(\opop \leq \idop)$,
 where $\idop$ denotes the identity operator;
 \emph{closings} $\clop$ that are increasing, idempotent, and extensive $(\clop \geq \idop)$.


%
%
A lattice operator $\psi$ is called \textit{decreasing}  or \textit{antitone}
if it is order-inverting, i.e.
$X\pord Y \Longrightarrow \psi(X)\ipord \psi (Y)$.
Dual homomorphisms
 interchange suprema with infima and hence are decreasing operators.
For example, \emph{anti-dilations} $\dlop^a$
satisfy  $\dlop^a(\bigvee _{i} X_i ) =\bigwedge _{i} \dlop^a (X_i)$.
A lattice dual automorphism is a bijection 
that interchanges suprema with infima. For example,
a \textit{negation}  $\nu$
is a dual automorphism that is also involutive, i.e. $\nu^2=\idop$.\\


\noindent
\textbf{Residuation and Adjunctions}:\

An increasing operator $\psi: \Ll \rightarrow \Mm$ between two complete lattices  is called \textit{residuated}  \cite{BlJa72,Blyt05}
if there exists an increasing operator $\eresid{\psi}: \Mm \rightarrow \Ll$ such that
\beq
\psi \eresid{\psi} \pord \idop \pord \eresid{\psi} \psi
\label{resopcl}
\eeq
$\eresid{\psi}$ is called the \textbf{residual} of $\psi$, is unique, and is the closest to being an inverse of $\psi$.
Specifically, the residuation pair $(\psi, \eresid{\psi})$ can solve inverse problems
of the type $\psi (X)= Y$ either exactly since  $\hat{X}= \eresid{\psi} (Y)$
is the greatest solution of $\psi (X)= Y$ if a solution exists,
 or approximately since $\hat{X}$ is the  \emph{greatest subsolution} in the sense that
\beq
\hat{X}= \eresid{\psi} (Y)=\bigvee \{ X:\psi (X)\pord Y \}
\label{resmaxsubsol}
\eeq
On complete lattices an increasing operator $\psi$ is residuated (resp. a residual $\eresid{\psi}$)
if and only if it is a dilation (resp. erosion).
Equivalently, $\psi$ is residuated if $\eresid{\psi} (Y)$,
defined as in (\ref{resmaxsubsol}), exists for each $Y$.
The residuation theory has been used for solving inverse problems (mainly in  matrix algebra)
  over the extended max-plus semiring $(\EREAL, \vee,+)$  or other complete idempotent semirings which as lattices are made complete \cite{Cuni76,Cuni79,BCOQ01,CGQ04}.

A pair $(\dlop,\erop)$  of two operators $\dlop: \Ll \rightarrow \Mm$ and $\erop: \Mm \rightarrow \Ll$
between two complete lattices  is called
\textbf{adjunction}\footnote{ As explained in \cite{BaBa93,Heij94,HeRo90},
the  adjunction is related to 
\emph{Galois connection}, which is a pair of two
decreasing maps $\psi$ and $\phi$ between complete lattices, such that $Y\leq \psi (X)$ $\Longleftrightarrow$ $X\leq \phi (Y)$; if this holds, both $\psi$ and $\phi$ are anti-dilations and
their compositions $\psi \phi$ and $\psi \phi$ are closings  \cite{Acha82}.
As a name, `adjunction' was introduced in \cite{Gie+80} as equivalent to an isotone Galois connection.
The advantage of residuations and adjunctions over  Galois connections is that the former can form new adjunctions via composition, whereas this is not the case with (antitone) Galois connections.
Several authors 
define residuations $(\psi, \eresid{\psi})$ on partially ordered sets;
this however may not guarantee the general existence of (\ref{resmaxsubsol}) and the fact that the
$(\psi, \eresid{\psi})$ are a dilation and erosion respectively, unless the underlying sets become complete lattices.
}
 if
\beq
\dlop (X)\leq Y \Longleftrightarrow X \leq \erop (Y)
\quad
\forall X\in \Ll, Y\in \Mm
\label{adjunction}
\eeq
In any adjunction,  $\dlop$ is a dilation and $\erop$ is an erosion.
The double inequality (\ref{adjunction}) is equivalent to the inequality (\ref{resopcl})
satisfied by a residuation pair of increasing operators
if we identify the residuated map $\psi$ with $\dlop$ and its residual $\eresid{\psi}$ with $\erop$.
Further, from (\ref{adjunction}) or (\ref{resopcl})  it follows that any adjunction $(\dlop, \erop)$ automatically yields  an opening $\opop=\dlop \erop$ and a closing $\clop=\erop \dlop$,
where the composition of two operators is written as an operator product.
To view $(\dlop,\erop)$ as an adjunction instead of a residuation pair has
the advantage of the additional  geometrical intuition and visualization afforded
by the dilation and erosion operators in image and shape analysis.

There is a one-to-one correspondence between the two operators
of an adjunction; e.g.,
given a dilation $\dlop$, there is a unique erosion
\beq
\erop (Y) =\eresid{\dlop}(Y)= \bigvee \{ X\in \Ll : \dlop (X)\pord Y\}
\eeq
such that $(\dlop,\erop)$ is an adjunction, and conversely.
Thus, dilations and erosions on complete lattices always come in pairs.
In any adjunction $(\dlop,\erop)$, $\erop$ is called the
\emph{adjoint erosion} (a.k.a. \emph{upper adjoint}) of $\dlop$, whereas $\dlop$ is
the \emph{adjoint dilation} (a.k.a. \emph{lower adjoint}) of $\erop$.

\begin{Example} {\rm
(a)~A classic example of a morphological set adjunction is the pair of Minkowski set addition $\dilt$ and subtraction $\eros$: for $X,B\sbs \REAL^n$
\beq
\begin{array}{rcl}
\dlop_B(X) = X\dilt B & \defineql & \{ \vct x\in \REAL^n: \tran{\refl{B}}{+\vct x} \cap X\neq \keno \}
\\
\quad \erop_B(X)=X\eros B & \defineql & \{ \vct x\in \REAL^n: \tran{B}{+\vct x} \sbs X \}
\end{array}
\eeq
where $\refl{B}=\{ -\vct b: \vct b \in B\}$ and $\tran{B}{+\vct x}\defineql \{ \vct b+\vct x: \vct b \in B\}$.\\
(b)~A classic example of a morphological signal adjunction is the pair of Minkowski function addition $\dilt$ and subtraction $\eros$: for $f,g: \REAL^n\rightarrow \EREAL$
\beq
\begin{array}{rcl}
\dlop_g(f)(\vct x)=f\dilt g (\vct x) & \defineql & \sup \{ f(\vct y-\vct x) +g(\vct y): \vct y \in \REAL^n \}
\\
\quad \erop_g(f)(\vct x)=f\eros g (\vct x) & \defineql & \inf \{ f(\vct x-\vct y) -g(\vct y): \vct y \in \REAL^n \}
\end{array}
\eeq
Thus, $f\dilt g$ is the supremal (max-plus) convolution of $f$ by $g$ and $f\eros g$ is the infimal convolution of $f(\vct x)$ by $-g(-\vct x)$.
} \end{Example}

%

\section{Tropical Dequantization and Vision Scale-Space PDEs}
\label{sc-deqntpde}

By following the procedure in Maslov \cite{Masl87},
we show in this section how the transformation of (\ref{lseapprox}) converts the classic linear heat PDE
to a nonlinear PDE that generates multiscale erosions (min-plus convolutions).
Consider the 1D linear heat PDE
\beq
\frac{\partial U}{\partial t} =\frac{\theta}{2}\frac{\partial^2 U}{\partial x^2}
\eeq
which models a homogeneous linear diffusion. It is also well known in computer vision because
it models  the Gaussian scale-space since its solution $U(x,t)$, $t\geq 0$, is the multiscale convolution of some initial function $f(x)=U(x,0)$ with multiscale Gaussians $G_\sigma(x)=\exp(-x^2/2\sigma^2)/(\sigma \sqrt{2\pi})$ of variance equal to $\sigma^2=\theta t$:
\beq
U(x,t) = \frac{1}{\sqrt{2\pi \theta t}}\int_{\REAL} f(x-y)\exp(-\frac{y^2}{2\theta t})dy
\label{heatpde}
\eeq
As shown by Maslov \cite{Masl87}, the substitution $U=\exp(-W/\theta)$ converts the heat PDE to  Hopf's nonlinear equation:
\beq
\frac{\partial W}{\partial t}+\frac{1}{2}\left(\frac{\partial W}{\partial x}\right)^2
-\frac{\theta}{2}\frac{\partial^2 W}{\partial x^2}=0
\label{hopfpde}
\eeq
The heat PDE (\ref{heatpde}) obeys a linear superposition; i.e., if $u_i(x,t)$ is its solution for initial condition $f_i(x)$, $i=1,2$, and if $f(x)=a_1f_1(x)+a_2f_2(x)$, the total solution becomes $U(x,t)=a_1u_1(x,t)+a_2u_2(x,t)$. However, the solution $W=-\theta\log U$ of the nonlinear PDE (\ref{hopfpde}) obeys the following nonlinear superposition:
\beq
W(x,t)=-\theta \log[\exp(-\frac{b_1+w_1}{\theta})+\exp(-\frac{b_2+w_2}{\theta})]
=-\theta \log[c_1u_1+c_2u_2]
\eeq
where $w_i(x,t)=-\theta\log u_i(x,t)$, $i=1,2$, are solutions of (\ref{hopfpde}) and $c_i=\exp(-b_i/\theta)$.
If the heat diffusivity constant $\theta$ becomes very small, we can perform Maslov's dequantization as in (\ref{lseapprox}) to convert the above log-sum-exp superposition to
a tropical (min-plus) superposition $W=\min(b_1+w_1,b_2+w_2)$ . 
Further, the limit of the PDE (\ref{hopfpde}) for $\theta\rightarrow 0$ yields
another nonlinear PDE
\beq
\frac{\partial S}{\partial t} +\frac{1}{2}\left(\frac{\partial S}{\partial x}\right)^2=0
\eeq
which models the multiscale weighted erosion $g(x)\eros k_t(x)$ of an initial function $g(x)=S(x,0)$ by multiscale parabolas $k_t(x)=-x^2/2t$:
\beq
S(x,t)=g(x)\eros k_t(x)=\bigwedge_y g(x-y)+y^2/2t
\eeq
 Thus, classic morphological PDEs \cite{BrMa94} can be obtained from the linear heat PDE (modeling Gaussian scale space) via tropicalization.

\section{Elements of Tropical Geometry}
\label{sc-elemtropgeo}

We first present some simple examples of tropical\footnote{\footnotesize The adjective `tropical' was coined by French mathematicians, including Dominique Perrin and Jean-Eric Pin, to honor their Brazilian colleague Imre Simon who was one of the pioneers of min-plus algebra as applied to automata. However, we give it an alternative and substantial meaning
 in connection with its Greek origin word \tg{``τροπικός"},
which comes from the Greek word \tg{``τροπή"} which means ``turn" or ``changing the way/direction",
to literally express the fact that tropical curves and surfaces bend and turn.} curves and surfaces which result from tropicalizing the polynomials that analytically describe their Euclidean counterparts; here `tropicalization' means replacing sum with max or min and multiplications with additions. Then, we explain this tropicalization as a dequantization of real algebraic geometry.

\subsection{Examples of Tropical Polynomial Curves and Surfaces}

\begin{figure}[!h]
\centering
\subfigure[Euclidean line]
{\includegraphics[width=0.3\textwidth]{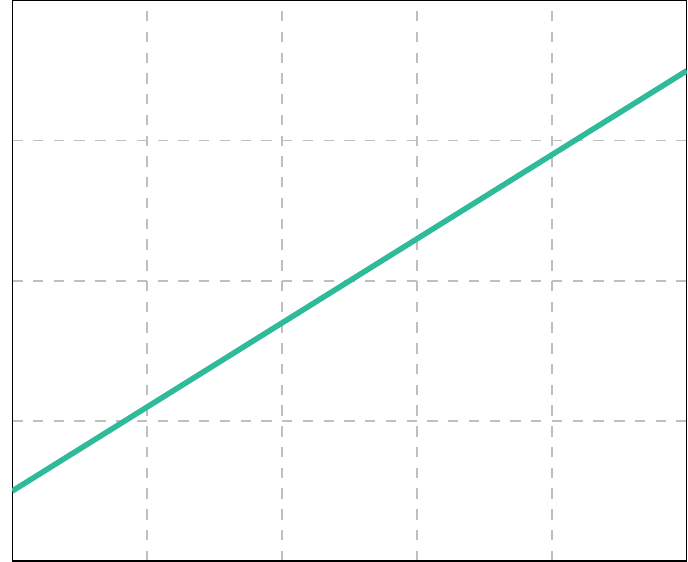}}
\subfigure[Euclidean parabola]
{\includegraphics[width=0.3\textwidth]{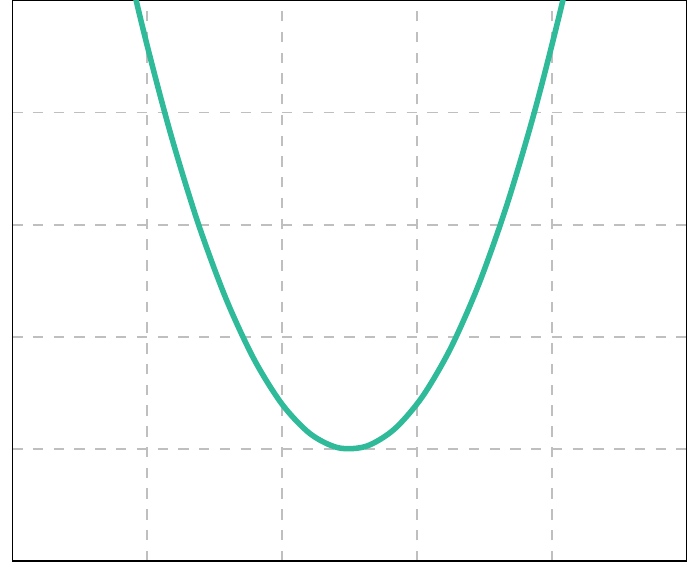}}
\subfigure[Euclidean cubic]
{\includegraphics[width=0.3\textwidth]{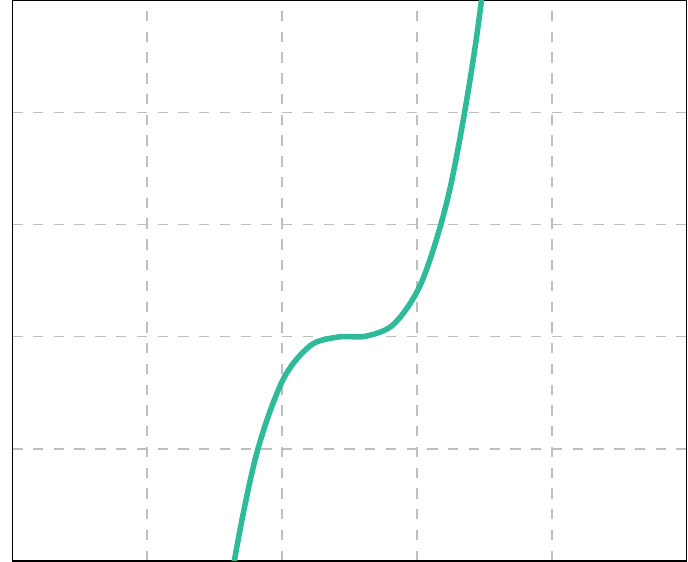}}
\\
\subfigure[Max-plus line]
{\includegraphics[width=0.3\textwidth]{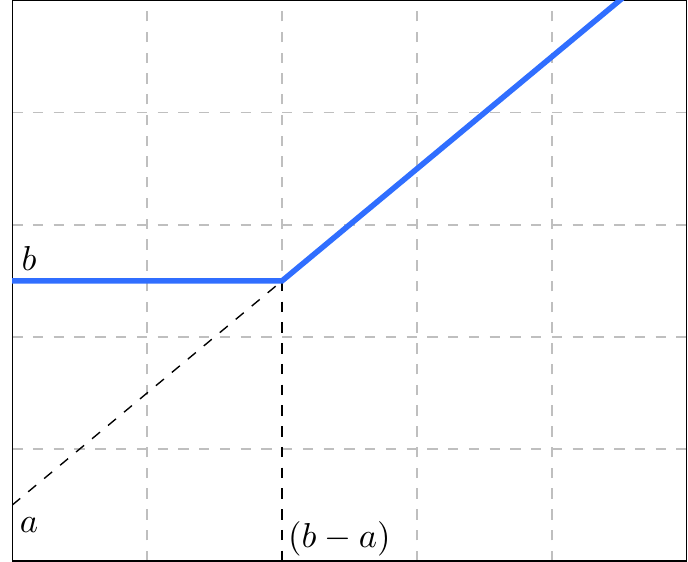}}
\subfigure[Max-plus parabola]
{\includegraphics[width=0.3\textwidth]{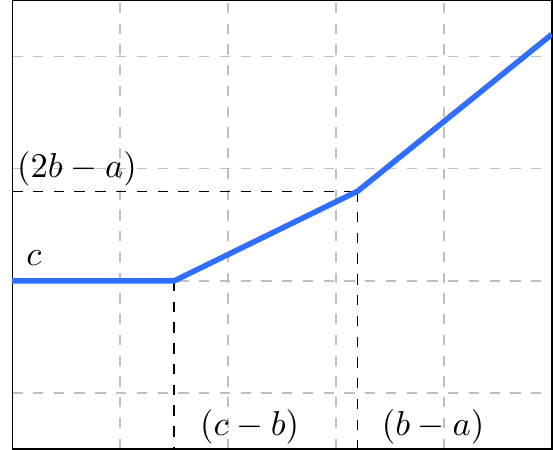}}
\subfigure[Max-plus cubic]
{\includegraphics[width=0.3\textwidth]{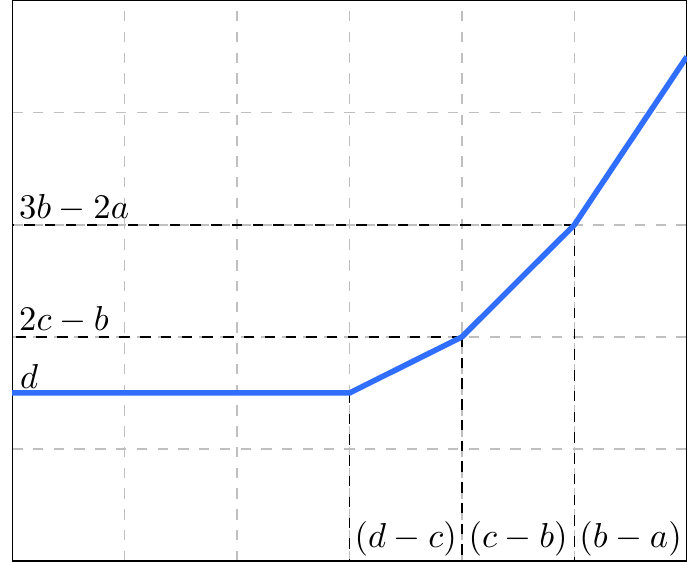}}
\\
\subfigure[Min-plus line]
{\includegraphics[width=0.3\textwidth]{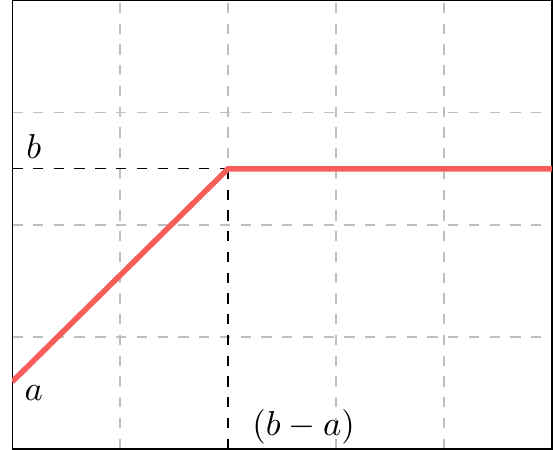}}
\subfigure[Min-plus parabola]
{\includegraphics[width=0.3\textwidth]{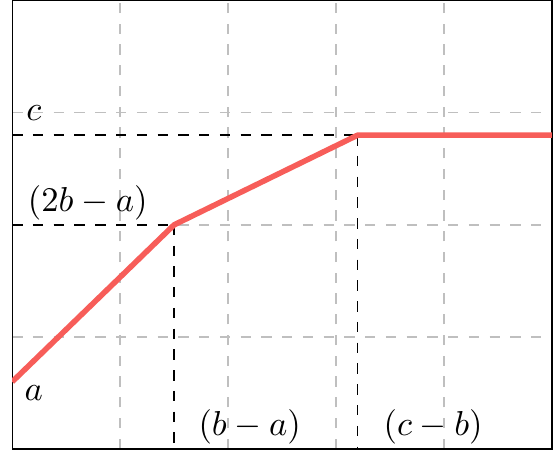}}
\subfigure[Min-plus cubic]
{\includegraphics[width=0.3\textwidth]{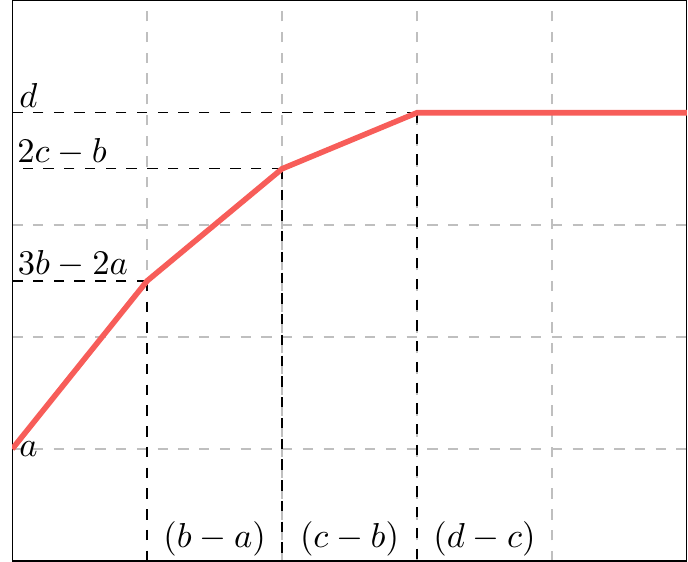}}
\caption{Euclidean and tropical 1D polynomials up to 3rd degree.}
\label{fg-euclid-tropic-polynom1}
\end{figure}

\noindent \textbf{Tropical Polynomial Curves}: \\
Consider the analytic expressions for a Euclidean line, parabola, and cubic curve:
\beq	
\begin{array}{lll}
		p_1(x) &= & a \cdot x + b, \quad
		p_2(x) =  a \cdot x^2 + b \cdot x + c,\\
		p_3(x) &= & a \cdot x^3 + b \cdot x^2 + c \cdot x + d,
\end{array}
\label{polynom1euclid_lpc}
\eeq
The equations for their corresponding max-plus tropical polynomials are:
\beq
\begin{array}{l}
		p^{\max}_1(x) = \max(a + x, b), \quad
		p^{\max}_2(x) = \max(a + 2\cdot x, b + x, c),\\
		p^{\max}_3(x) = \max(a + 3\cdot x, b + 2\cdot x, c + x, d),
\end{array}
\label{polynom1tropic_lpc}
\eeq
The equations for the min-plus case are identical as in (\ref{polynom1tropic_lpc})
by replacing max with min.
The graphs of all the above can be seen in Fig.~\ref{fg-euclid-tropic-polynom1}.
\\

\begin{figure}[!h]
		\begin{center}
\subfigure[Surface $f_1$ (max-plus)]
{\includegraphics[width=0.45\textwidth]{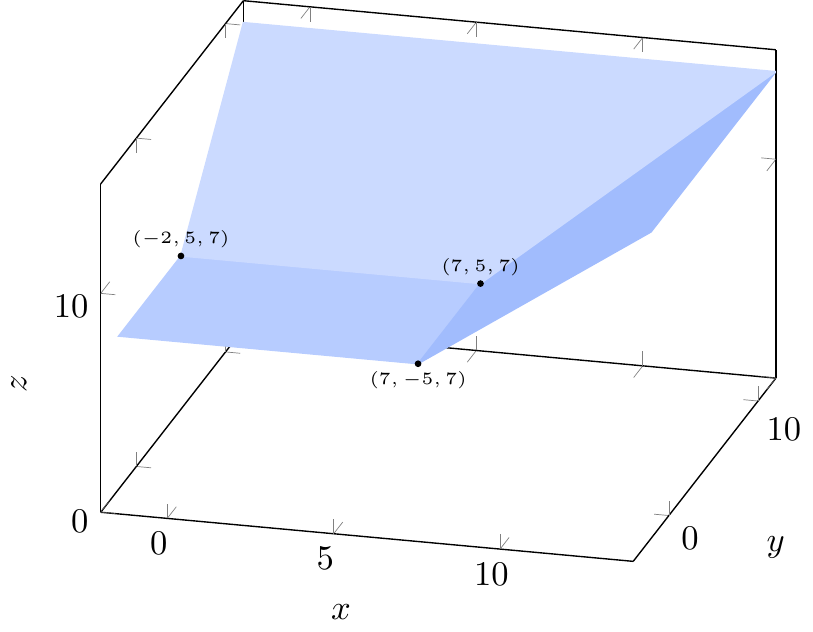}}
\subfigure[Surface $f_2$ (min-plus)]
{\includegraphics[width=0.45\textwidth]{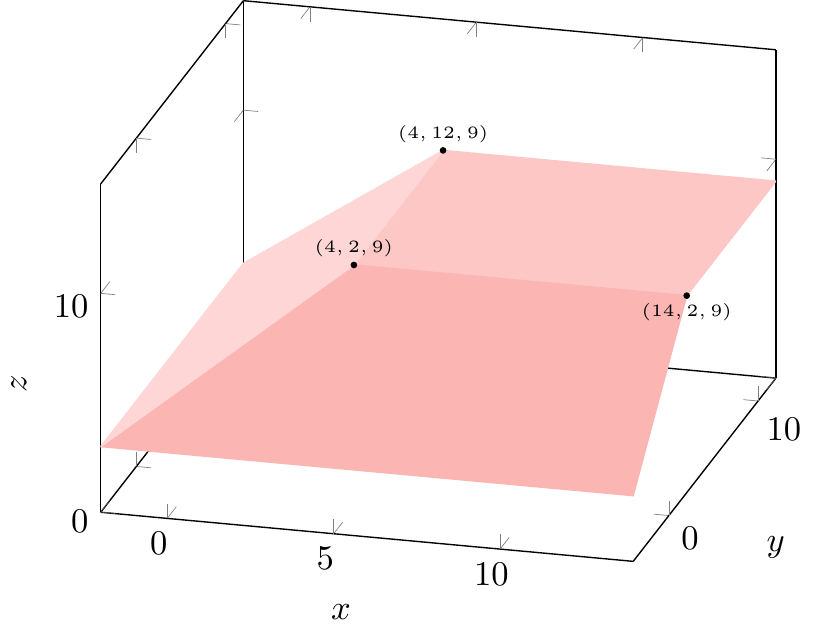}}
		\end{center}
		\caption{Surfaces of the two tropical planes  in (\ref{tropplane_min_max}).}
		\label{fg-tropplanes}
	\end{figure}

\noindent \textbf{Tropical Polynomial Surfaces}:\\
Consider the equations of the following tropical planes represented as 2D max-plus and min-plus polynomial of degree 1:
\beq
		f_1(x, y) = \max(0 + x, 2 + y, 7), \quad
		f_2(x, y) = \min(5 + x, 7 + y, 9),
\label{tropplane_min_max}
\eeq
whose graphs can be seen as surfaces in Fig.~\ref{fg-tropplanes}.

\begin{figure}[!h]
\centering
\includegraphics[width=0.6\columnwidth]{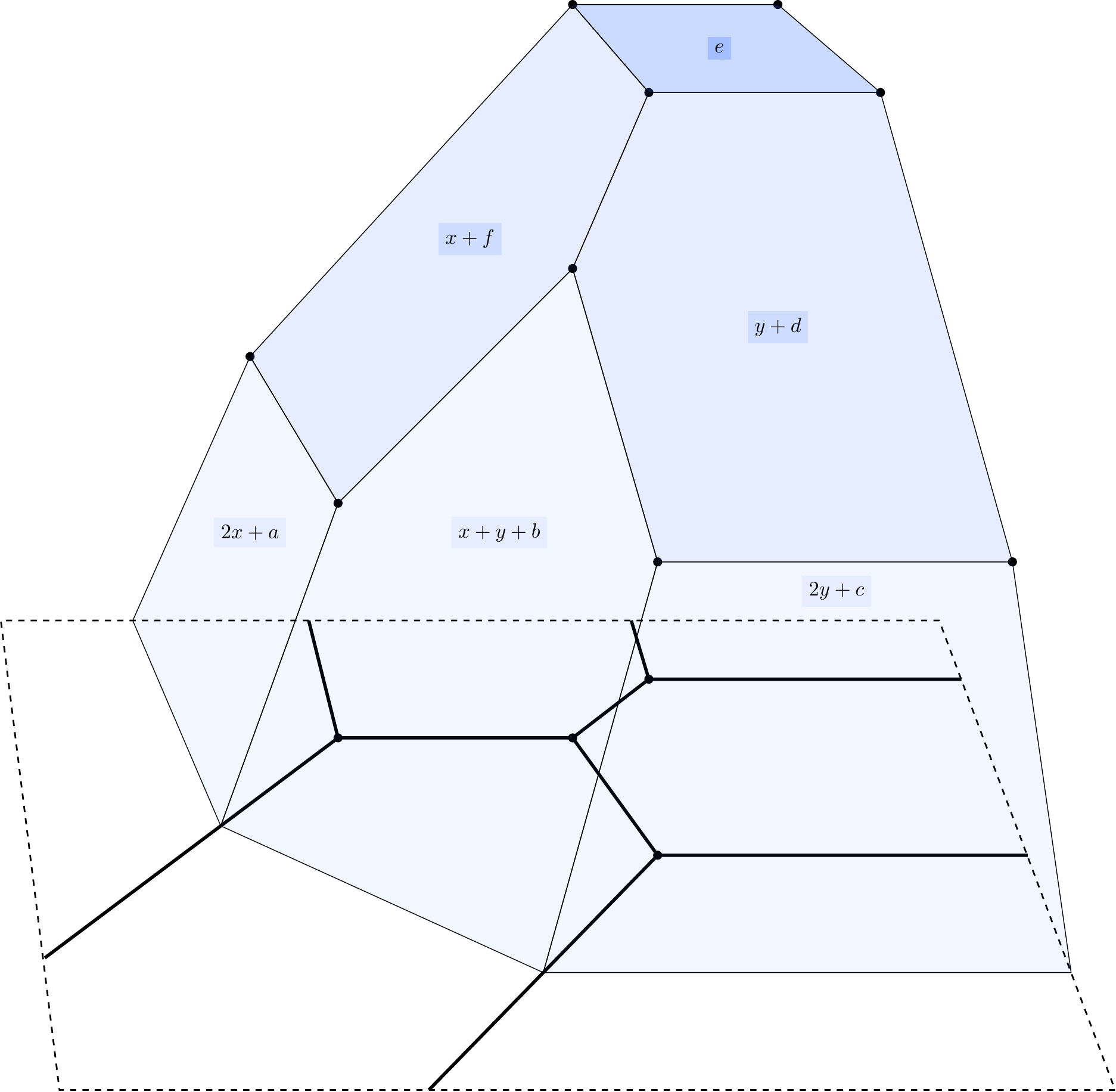}
\caption{Surface (graph) of the 2D min-plus tropical polynomial function $p(x,y)=\min(a+2x,b+x+y,c+2y,d+x,e+y,f)$ and its tropical quadratic curve. (Inspired by Fig.~1.3.2 of \cite{MaSt15}.)}
\label{fg-minsumpolynom2}
\end{figure}

As a next example, to the general Euclidean conic polynomial
\beq
p_{\textrm{e-conic}}(x,y)=ax^2+bxy+cy^2+dx+ey+f
\eeq
there corresponds the following two-variable max-plus tropical polynomial of degree 2:
\beq
p_{\textrm{t-conic}}(x,y)=\max(a+2x,b+x+y,c+2y,d+x,e+y,f)
\label{tropconic}
\eeq
Its min-plus version is shown in Fig.~\ref{fg-minsumpolynom2}.

\subsection{Tropical Polynomials as Dequantization of Algebraic Geometry}

The algebraic side of tropical geometry \cite{MaSt15} results from a transformation of analytic Euclidean geometry where the traditional  arithmetic of the real field $(\REAL,+,\times)$ involved in the analytic expressions of geometric objects is replaced by the arithmetic of the min-plus tropical semiring $(\REAL_{\min},\wedge, +)$; some authors use its max-plus dual semiring $(\REAL_{\max},\vee, +)$.
We use both semirings as dual parts of the weighted lattice $(\EREAL, \vee,\wedge,+)$ (explained in Sec.~\ref{sc-wl}). This transformation converts Euclidean objects into polygonal lines on the plane and polyhedra in higher dimensions. A geometric explanation and visualization of this transformation is obtained
from Viro's graphing of polynomial curves on log-log paper \cite{Viro01}. Consider the monomial curve $v=cu^a$, $c>0$,  on the positive quadrant
of the $(u,v)$ plane and consider the log-log transformation of both coordinates composed with a uniform scaling by $\theta >0$: $x=\theta\log u$, $y=\theta\log v$. Then, on the $(x,y)$ plane the curve becomes the line $y=b/\theta+a x$, where $b=\log c$.
If we have a $K$-term polynomial curve $v=P(u)=\sum_{k=1}^Kc_ku^{a_k}$ with $c_k=\exp(b_k) > 0$ and $a_k\in \REAL$ (i.e. a posynomial \cite{BKVH07}) then
we convert it to
\beq
P_\theta(x)=\theta \log [\sum_{k=1}^K \exp(b_k/\theta)\exp (a_kx/\theta)]
\eeq
As $\theta\downarrow 0$ this  yields via Maslov dequantization a $K$-term \textbf{1D max-plus tropical polynomial}
\beq
p(x)=\max_{k=1}^K [b_k+a_kx]
\eeq
While each $P_\theta(x)$ is a smooth function, their limit $p(x)$ is a max-affine function and represents a \emph{piecewise-linear (PWL)} convex function. Note: if we perform dequantization with negative exponents as in (\ref{lseapprox})  we obtain a min-plus polynomial which is a PWL concave function.

The above procedure extends to multiple dimensions or higher degrees and shows us the way to tropicalize any classic $n$-variable polynomial (linear combination of power monomials) $\sum _k c_k u_1^{a_{k1}}\cdots u_n^{a_{kn}}$ defined over $\REAL_{>0}^n$ where $c_k> 0$ and $\vct a_k=(a_{k1},...,a_{kn})^T$ is traditionally some nonnegative integer\footnote{Traditionally, `tropical polynomials' assume that the parameters $a_{ki}$ are nonnegative integers. If we also allow negative integers, we get `Laurent tropical polynomials'. As in \cite{Butk10}, we allow any real coefficients; this may be called  `tropical posynomials' \cite{CGP19a}.}
vector but herein we allow $\vct a_k\in \REAL^n$: replace the sum with max and log the individual terms so that the multiplicative coefficients become additive and the powers become real multiples of the indefinite log variables. Thus, a general $n$-variable max-plus polynomial $p:\REAL^n\rightarrow \REAL$ has the expression:
\beq
p(\vct x)=\bigvee_{k=1}^K b_k+\vct a_k^T\vct x, \quad \vct x=(x_1,...x_n)^T
\label{mspolynom}
\eeq
where  $K=\rank(p)$ is the number of terms of $p$.
Its graph (hypersurface) is a max of $K$ hyperplanes with intercepts $b_k=\log c_k\in \REAL$ and real slope vectors $\vct a_k\in \REAL^n$.
The degree of $p$ is $|\vct a|=\max_k|\vct a_k|$ where $|\vct a_k|=|a_{k1}|+\cdots +|a_{kn}|$.
Thus, the curves or surfaces of real algebraic geometry, which is essentially polynomial geometry, become via dequantization the
graphs of \emph{convex} PWL functions represented by tropical (max-plus) polynomials.

\subsection{Tropical Curves and Newton Polytopes}

\begin{figure}[!h]
\centering
\subfigure[Max-plus curve]
{\includegraphics[width=0.3\columnwidth]{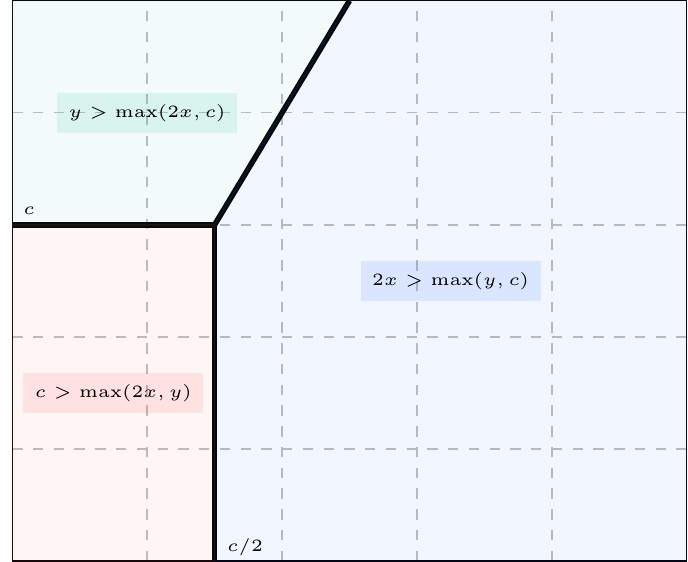}}
\hspace{5mm}
\subfigure[Min-plus curve]
{\includegraphics[width=0.3\columnwidth]{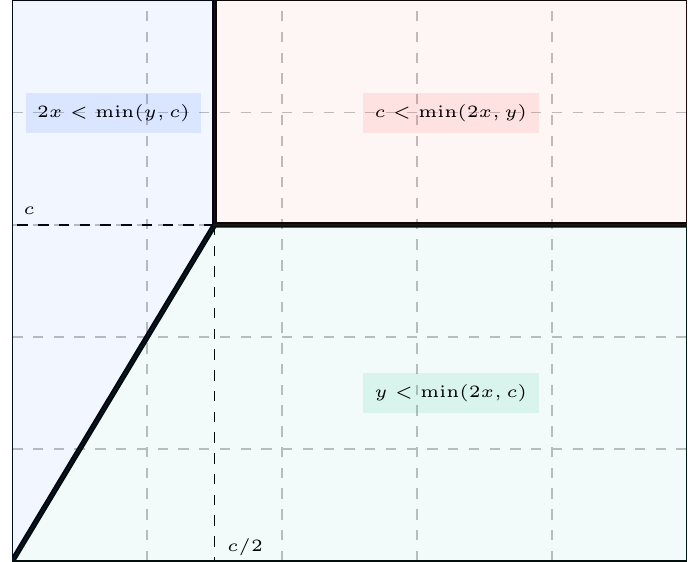}}
\caption{Tropical curve of the max-polynomial $p(x,y)=\max(2x,y,c)$ left and its dual min-polynomial $p'(x,y)=\min(2x,y,c)$ right. Best viewed in color.}
\label{fg-tropsurf}
\end{figure}

To the zero set of a classic polynomial there corresponds the \emph{tropical curve or surface}
of a max-plus tropic polynomial $p:\REAL^n\rightarrow \REAL$
\beq
\Vv (p) \defineq \{ \vct x\in \REAL^n:
 \mbox{\rm more than one terms of $p(\vct x)$ attain the max} \}
 \label{tropcurve}
\eeq
The above also defines the tropical curve of min-plus polynomials by replacing max with min.
Thus, $\Vv (p)$ consists of the singularity points (of non-differentiability) of $p(\vct x)$.
Examples are shown in Fig.~\ref{fg-tropsurf} for degree-1 tropic polynomials and in Fig.~\ref{fg-minsumpolynom2} for a degree-2 polynomial.
The max-plus line $y=\max (a+x,b)$ of Fig.~\ref{fg-euclid-tropic-polynom1} and the tropical curve of the max-plus polynomial
$\max (a+x,b+y,c)$ of Fig.~\ref{fg-tropsurf} are special cases of the general family of the 12 max-plus line types of $\Rmax^2$ given
in \cite{CGQ04} by
\beq
\max (a+x,b+y,c)=\max (a'+x,b'+y,c'), \quad a,b,c,a',b',c' \in \Rmax,
\eeq
where not all the coefficients are needed. This is a tropical version of the Euclidean line equation $ax+by+c=0$.

\begin{figure}[!h]
\centering
\subfigure[Polytopes]
{\includegraphics[width=0.3\columnwidth]{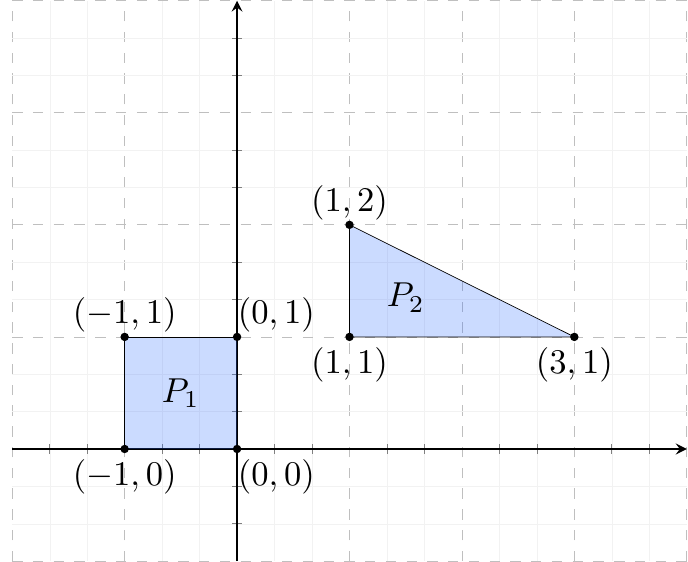}}
\hspace{3mm}
\subfigure[Newton (max)]
{\includegraphics[width=0.3\columnwidth]{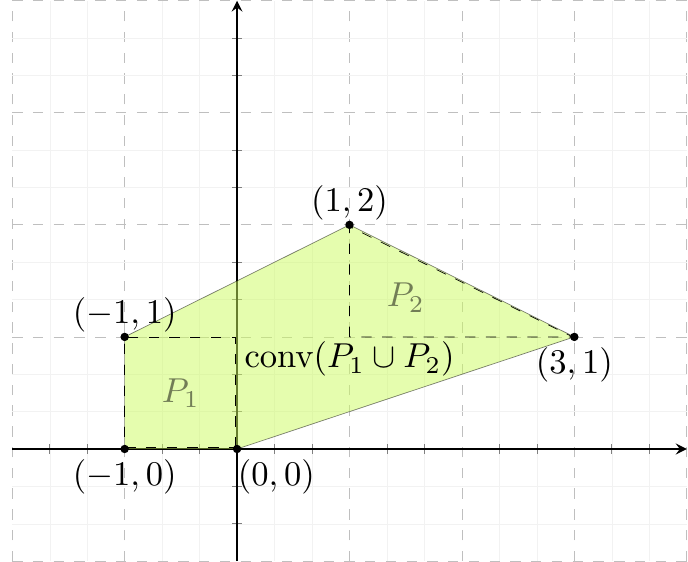}}
\hspace{3mm}
\subfigure[Newton (sum)]
{\includegraphics[width=0.3\columnwidth]{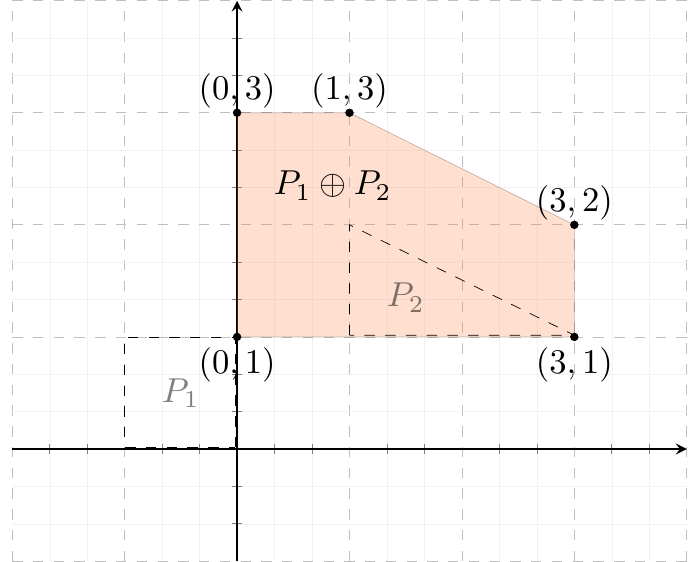}}
%
\caption{Newton polytopes of (a)~two max-polynomials $p_1(x,y)=\max(x+y,3x+y,x+2y)$ and $p_2(x,y)=\max(0,-x,y,y-x)$, (b)~their max $p_1 \vee p_2$, and (c)~their sum $p_1 + p_2$.}
\label{fg-newton}
\end{figure}

Another interesting geometric object related to a max-plus polynomial $p$ is its \emph{Newton polytope}
which is the convex hull (denoted by $\conv(\cdot )$) of the set of points represented by its slope coefficient vectors:
\beq
\newtpoly (p) \defineq \conv \{ \vct a_k: k=1,...,\rank(p)\}
\eeq
This satisfies several important properties \cite{ChMa17}:
\begin{eqnarray}
\newtpoly (p_1 \vee p_2) & = & \conv (\newtpoly(p_1)\cup \newtpoly(p_2)) \\
\newtpoly (p_1 + p_2) & = & \newtpoly(p_1)\dilt \newtpoly(p_2)
\end{eqnarray}
Examples are shown in Fig.~\ref{fg-newton}.
Thus, the Newton polytope of the sum (resp. max) of two tropical polynomials is the Minkowski sum (resp. the convex hull of the union) of their individual polytopes.

\subsection{Tropical Halfspaces and Polytopes}

\begin{figure}[!h]
\centering
\subfigure[Single region]
{\includegraphics[width=0.35\columnwidth]{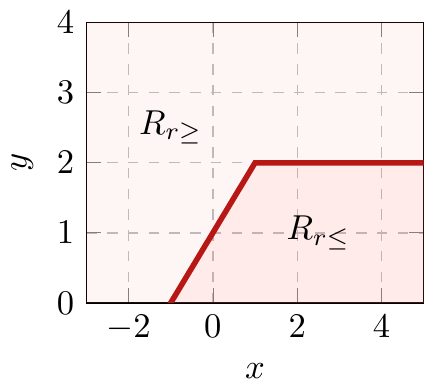}}
\hspace{5mm}
\subfigure[Multiple regions]
{\includegraphics[width=0.35\columnwidth]{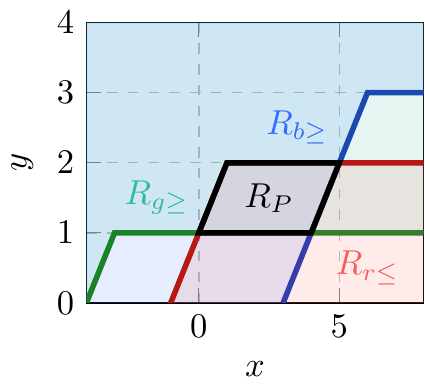}}
\caption{Regions $R_{c\geq}$ and $R_{c\leq}$ formed by min-plus tropical halfspaces in $\REAL^2$, where $c$ denotes the color of the tropical boundary and $\geq 0$ (resp. $\leq 0$) the set of points above (resp. below) the boundary. (a)~The red boundary is the min-plus tropical line $y=\min (1+x,2)$. (b)~The green and blue boundaries are respectively the  tropical lines $y=\min (4+x,1)$ and $y=\min (x-3,3)$. $R_P$ is the polytope formed by the intersection of three tropical halfplanes.
Best viewed in color.}
\label{fg-t-halfsp2}
\end{figure}

In pattern analysis problems on Euclidean spaces $\REAL^{n+1}$ we often use halfspaces
$\Hh (\vct a,b)\defineql \{ \vct x\in \REAL^n: \vct a^T\vct x\leq b\}$, polyhedra (finite intersections of halfspaces),
 and polytopes (compact polyhedra formed as the convex hull  of a finite set of points).
%
%
%
Replacing linear inner products $\vct a^T\vct x$ with max-plus versions yields \emph{tropical halfspaces} \cite{GaKa11} with parameters $\vct a=[a_i],\vct b=[b_i]\in \REAL^{n+1}$:
\beq
\Tt (\vct a,\vct b)\defineq \{  \vct x\in \REAL_{\textrm{max}}^n:
\max( a_{n+1}, \bigvee_{i=1}^n a_i+x_i) \leq
\max( b_{n+1}, \bigvee_{i=1}^n b_i+x_i) \}
\label{max+halfsp}
\eeq
where $\min(a_i,b_i)=-\infty$ $\forall i$.
Thus, for each $i$, only one coefficient is needed either in the left or in the right side of inequality (\ref{max+halfsp}).
Replacing max with min yields tropical halfspaces with dual boundaries that are min-plus hyperplanes.
Examples of polytopes in the plane that are polygonal regions formed by  min-plus tropical halfplanes are shown in Fig.~\ref{fg-t-halfsp2}.
Obviously, their separating boundaries are tropical lines. Such regions in multiple dimensions were used in \cite{ChMa17,ChMa18,YaMa95} as morphological perceptrons.

\begin{figure}[!h]
		\begin{center}
\subfigure[First view]
{\includegraphics[width=0.45\textwidth]{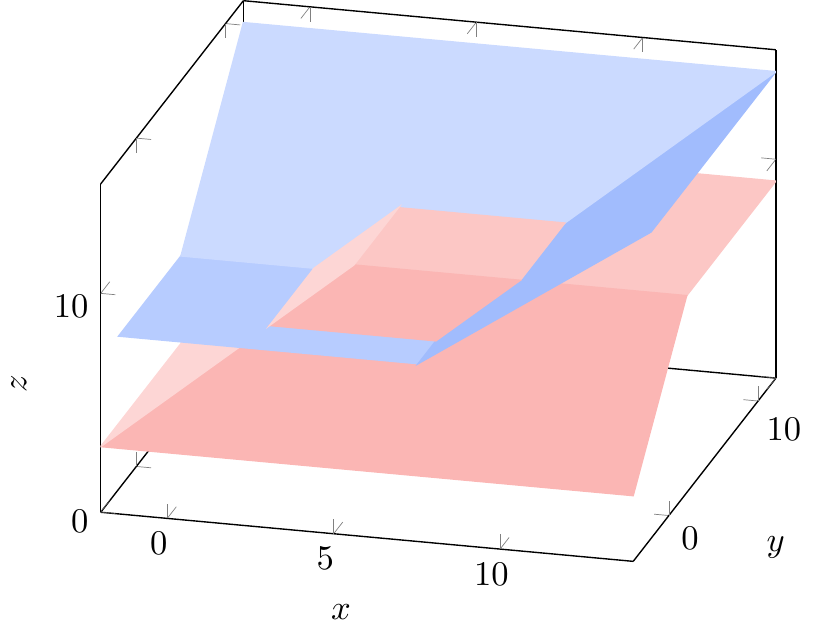}}
\subfigure[Second view]
{\includegraphics[width=0.45\textwidth]{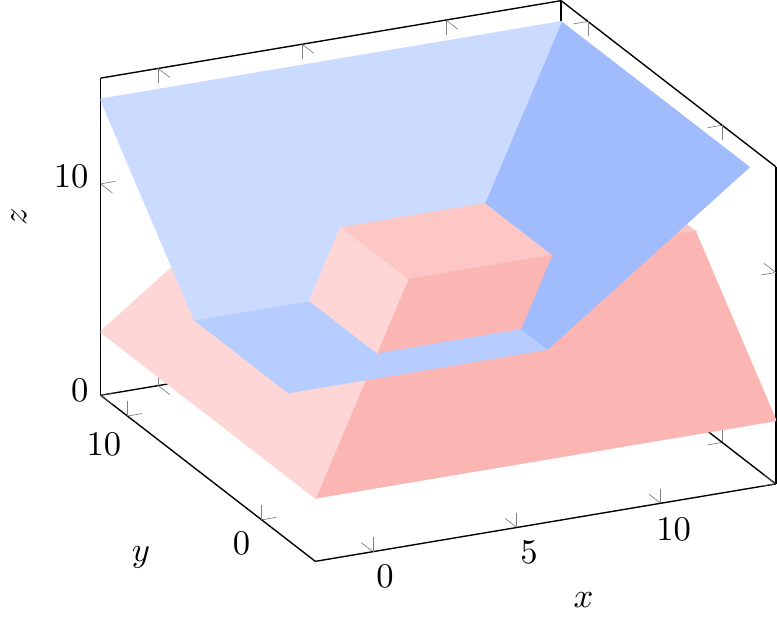}}
		\end{center}
\caption{Intersection of halfspaces of the 2D max-plus and min-plus tropical polynomials in (\ref{tropplane_min_max}). Best viewed in color.}
	\label{fg-polyhedron}
\end{figure}

As an example in 3D space, in Fig.~\ref{fg-polyhedron} we can see two different views of the intersection of the tropical halfspaces corresponding to the two tropical polynomial in (\ref{tropplane_min_max}). This is a polytope that is the polyhedral region formed by intersecting the halfspace above the surface of the 2D max-plus polynomial $f_1$ with the halfspace below the surface of the min-plus polynomial $f_2$.

We note from Fig.~\ref{fg-t-halfsp2} and Fig.~\ref{fg-polyhedron} that the number of tropical boundaries required to form polytopes,
which could serve as decision regions in pattern classification problems,
 is smaller than the number of linear boundaries.
 See, for instance, the polytope $R_P$ in Fig.~\ref{fg-t-halfsp2}(b). This observation remains valid in higher dimensions too;
 namely, decision regions can be formed with fewer tropical lines or hyper-planes (and hence with fewer parameters) than their Euclidean counterparts because any single tropical line or hyper-plane bends and turns and hence provides more than one straight line edge or flat face.

\section{Weighted Lattices: Nonlinear Vector Spaces and Extensions of Tropical Algebra and Geometry}
\label{sc-wl}

\subsection{Clodum: Extending Tropical Scalar Arithmetic}
\label{sc-lomclodum}


A lattice $(\Kk,\vee,\wedge)$ is often endowed with a third binary operation,
called symbolically the `multiplication' $\mgop$, under which
$(\Kk,\mgop)$ is  a group, or a monoid, or just a semigroup \cite{Birk67}.
Even if we have only a sup-semilattice $(\Kk,\vee)$ (i.e. an idempotent commutative semigroup) we can consider its supremum $\vee$ as an idempotent `addition' and equip it with an additional `multiplication' operation $\mgop$ so that the structure $(\Kk,\vee,\mgop)$ becomes an idempotent semiring.
Such ordered monoids have been studied in detail in \cite{Birk67,Zimm81,GoMi08} and form the algebraic basis
of max-plus algebra.

Consider now an algebra $(\Kk,\vee,\wedge,\mgop,\dmgop)$ with four binary operations,
which we call a \emph{lattice-ordered double monoid},
where $(\Kk,\vee,\wedge)$ is a lattice,
$(\Kk,\mgop)$ is a monoid whose `multiplication' $\mgop$ distributes over $\vee$,
and $(\Kk,\dmgop)$ is a  monoid whose `multiplication' $\dmgop$ distributes over $\wedge$.
These distributivities imply that both $\mgop$ and $\dmgop$ are increasing.
To the above definitions we add the word \emph{complete}
if $\Kk$  is a complete lattice and the distributivities involved are infinite.
We call the resulting algebra 
a  \emph{complete lattice-ordered double monoid}, in short \textit{clodum} 
 \cite{Mara05a,Mara13,Mara17}.
Previous works on minimax or max-plus algebra
have  used alternative names
for   structures similar to the above definitions
which emphasize semigroups and semirings instead of lattices
\cite{BCOQ01,Cuni79,GoMi08}; see \cite{Mara17} for similarities and differences.
%
We precisely define an algebraic structure $(\Kk, \vee ,\wedge, \mgop,\dmgop )$ to be a \textbf{clodum} if: \\
(C1)~$(\Kk,\vee,\wedge)$ is a complete distributive lattice.
Thus,
it contains its least $\vsetle \defineql \bigwedge \vset$
  and greatest element $\vsetge \defineql \bigvee \vset$.
The  supremum $\vee$ (resp. infimum $\wedge$) plays the role of a generalized \textit{`addition'} (resp. \textit{`dual addition'}). \\
(C2)~$(\Kk,\mgop)$ is a monoid  whose operation $\mgop$ plays the role of a generalized \textit{`multiplication'} with identity (`unit') element $\mgid$ and is a dilation. \\
(C3)~$(\Kk,\dmgop)$ is a monoid with identity  $\dmgid$ whose operation $\dmgop$ plays the role of a generalized \textit{`dual multiplication'} and is an erosion.

Remarks:
(i)~As a lattice, $\Kk$ is not necessarily infinitely distributive, although herein all our examples will be such.  \\
(ii)~The clodum `multiplications' $\mgop$ and $\dmgop$ do not have to be commutative. \\
(iii)~The least (greatest) element $\ltle$ ($\ltge$) of $\Kk$ is both the `zero' element
for the `addition' $\vee$ ($\wedge$) and an absorbing null for the `multiplication' $\mgop$ ($\dmgop$). \\
(iv)~We avoid degenerate cases by  assuming that
 $\vee \not = \mgop$ and $\wedge \not = \dmgop$.
However,  $\mgop$ may be the same as $\dmgop$, in which case
we have a self-dual `multiplication'. \\

%
%
A clodum $\vset$ is called \emph{self-conjugate}
if  it has a lattice negation  $a\mapsto \glconj{a}$
such that
\beq
\glconj{( \bigvee_i a_i )}  =  \bigwedge_i \glconj{a_i} \; \; , \; \;
\glconj{( \bigwedge_i b_i )} =  \bigvee_i \glconj{b_i}\; \; , \; \;
\glconj{(a\mgop b)}  =  \glconj{a}\dmgop \glconj{b}
\label{clodconj}
\eeq
The first two above properties are generalization of De Morgan's laws in Boolean algebras.
We assume that the suprema and infima in (\ref{clodconj}) may be over any (possibly infinite) collections.

%
%

If  $\mgop=\dmgop$
over $G=\vset \setminus \{ \ltle,\ltge\}$ where $(G,\mgop)$ is a group and $(G,\vee,\wedge)$
a conditionally complete lattice,
then the clodum $\vset$ becomes a richer structure
which we call a \emph{complete lattice-ordered group}, in short \textbf{clog}.
 In any clog the distributivity between $\vee$ and $\wedge$ is of
the infinite type and the `multiplications' $\mgop$ and $\dmgop$ are commutative.
%
Then, for each $a\in \vsgroup$ there exists its `multiplicative inverse'
$a^{-1}$ such that $a\mgop a^{-1}=\mgid $.
%
Further, the `multiplication' $\mgop$ and its self-dual
$\dmgop$ (which coincide over $\vsgroup$) can be extended over the whole
$\vset$ by
involving the null elements.
%
A clog becomes self-conjugate  by setting
$\glconj{a}= a^{-1}$ if $\vsetle \spord a \spord \vsetge$,
$\glconj{\vsetge}=\vsetle$, and $\glconj{\vsetle}=\vsetge$.
In a clog $\vset$ the $\mgop$ and $\dmgop$ coincide in all cases with only one exception:
the combination of the least and greatest elements;
thus, we may occasionally denote the clog algebra as $(\vset, \vee,\wedge,\mgop)$.
%

\begin{Example}\ \label{ex-cloda} {\rm
(a)~\emph{Max-plus} clog $(\EREAL,\vee,\wedge,+,+')$:\
$\vee/\wedge$ denote the standard sup/inf on $\EREAL$,
$+$ is the standard addition on  $\EREAL$  playing the role of a `multiplication' $\mgop$
with $+'$ being the `dual multiplication' $\dmgop$;
 the operations $+$ and $+'$ are identical for finite reals, but $a+(-\infty)=-\infty$
and $a+'(+\infty)=+\infty$ for all $a\in \EREAL$.
The identities are $\mgid=\dmgid=0$, the nulls are $\vsetle=-\infty$ and $\vsetge=+\infty$, and the
conjugation mapping is $\glconj{a}=-a$.
Thus, $+$ and $+'$ are respectively the `lower addition' and `upper addition' used  in convex analysis \cite{More70}.
\\
(b)~\emph{Max-times} clog $([0,+\infty],\vee,\wedge,\times,\times')$:
The identities are $\mgid=\dmgid=1$, the nulls are $\vsetle=0$ and $\vsetge=+\infty$, and the
conjugation mapping is $\glconj{a}=1/a$.
The scalar multiplications $\times$ and $\times'$ coincide over $(0,+\infty)$,
but $a\times 0=0$ and $a\times '(+\infty) =+\infty$ for all $a\in [0,+\infty]$.
\\
(c)~\emph{Max-min} clodum \ $([0,1],\vee,\wedge,\min,\max)$:
As `multiplications' we have $\mgop=\min$ and $\dmgop=\max$.
The identities and nulls are $\dmgid=\vsetle=0$, $\mgid=\vsetge=1$.
A possible conjugation mapping is $\glconj{a}=1-a$.
Additional clodums that are not clogs are discussed in \cite{Mara05a,Mara17} using
more general fuzzy intersections and unions.
\\
(d)~\emph{Max-softmin} clodum \ $(\EREAL,\vee,\wedge,+_{-\theta},+_\theta)$, $\theta>0$:
As `multiplication' we have  $\mgop=+_{-\theta}$ and as `dual multiplication'  $\dmgop=+_\theta$,
defined in the log-sum-exp approximation (\ref{lsesemiring}).
The identities and nulls are $\dmgid=\vsetle=-\infty$, $\mgid=\vsetge=+\infty$, and the
conjugation mapping is $\glconj{a}=-a$. By varying $\theta>0$ we obtain a family of
clodums whose `multiplications' $\mgop$ and $\dmgop$ are smooth (`soft') versions of the min and max operations respectively.
In the limit as $\theta\downarrow 0$ this family converges to a max-min clodum over $\EREAL$.
\\
(e)~\emph{Matrix} max-$\mgop$ clodum: $(\vset^{n\times n},\vee,\wedge,\mxgmp,\mngmp)$
where $\vset^{n\times n}$ is the set of $n\times n$ matrices with entries from a clodum $\vset$,
$\vee$/$\wedge$ denote here elementwise matrix sup/inf,
and $\mxgmp,\mngmp$ denote max-$\mgop$ and min-$\dmgop$ matrix `multiplications':
\[
\mtr{C} = \mtr{A} \mxgmp \mtr{B}=[c_{ij}],
 \; c_{ij} = \bigvee _{k=1}^n a_{ik}\mgop b_{kj}
\; , \;
\mtr{D} = \mtr{A} \mngmp \mtr{B}=[d_{ij}], \;
  d_{ij} = \bigwedge _{k=1}^n a_{ik}\dmgop b_{kj}
\label{maxmingenmpr}
\]
This is a clodum with non-commutative  `multiplications'.
For the max-plus clog $(\EREAL,\vee,\wedge,+,+')$, these matrix `multiplications' are defined and denoted as
\beq
[\mtr{A} \mxsmp \mtr{B}]_{ij}\defineq  \bigvee  _{k=1}^n a_{ik} + b_{kj}
\; , \;
[\mtr{A} \mnsmp \mtr{B}]_{ij}\defineq \bigwedge _{k=1}^n a_{ik} +' b_{kj}
\label{maxminsummpr}
\eeq
} \end{Example}

\subsection{Complete Weighted Lattices: Nonlinear Vector Spaces}
\label{sc-cwl}
Consider a nonempty collection $\Ww$ of mathematical objects, which will be our
space; examples of such objects include the vectors in $\EREAL^n$ or signals in
$\fun(E,\EREAL)$.
Also, consider a
clodum $(\vset, \vee, \wedge, \mgop, \dmgop )$ of \textbf{scalars}
with \emph{commutative}  operations $\mgop,\dmgop$ and $\vset \sbs \EREAL$.
%
%
We define  \emph{two internal operations} among vectors/signals $X,Y$ in $\Ww$:
their supremum $X\vee Y:\Ww^2\rightarrow \Ww$ and their infimum $X\wedge Y:\Ww^2\rightarrow \Ww$,
which we denote using the same supremum symbol ($\vee$) and infimum symbol ($\wedge$) as
in the clodum, hoping that the differences will be clear to the reader from the
context. Further, we define \emph{two external operations} among any vector/signal $X$ in $\Ww$
and any scalar $c$ in $\vset$:
a `scalar multiplication' $c\mgop X:(\vset,\Ww)\rightarrow \Ww$ and
a `scalar dual multiplication' $c\dmgop X:(\vset,\Ww)\rightarrow \Ww$,
again by using the same symbols as in the clodum.
 Now, we define $\Ww$ to be a \textbf{weighted lattice} space
 over the clodum $\vset$ if for all $X,Y,Z\in \Ww$ and  $a,b\in \vset$ all the axioms of
Table~\ref{tb-wlat} hold.
Note that, 
under axioms L1-L9 and their duals L1$'$-L9$'$,
$\Ww$ is a distributive lattice with a least element $\flatle$ and a greatest element $\flatge$.
These  axioms bear a striking similarity with those of a linear space.
One difference is that the vector/signal addition ($+$) of linear spaces
is now replaced by two dual superpositions,
the lattice supremum ($\vee$) and infimum ($\wedge$);
further, the scalar multiplication ($\times$) of linear spaces is now replaced
by two  operations $\mgop$ and $\dmgop$ which are dual to each other.
Only one major property of linear
spaces is missing from the  weighted lattices: the existence of `additive inverses'.
%
%
%
We define the space $\Ww$ 
to be a \textbf{complete weighted lattice (CWL)}
 if
 (i)~\ $\Ww$ is closed under any (possibly infinite) suprema and infima, and
(ii)~the distributivity laws between the scalar operations  $\mgop$ ($\dmgop$)
and the supremum (infimum)  are of the infinite type.
Note that a commutative clodum is a  complete weighted lattice over itself.

\begin{table*}[!h]
\caption{Axioms of Weighted Lattices  \protect{\cite{Mara17}}}
\begin{tabularx}{\textwidth}{|X|X|X|} \hline
Sup-Semilattice & Inf-Semilattice & Description \\ \hline \hline
L1. $\; X\vee Y\in \Ww$ & L1$'. \; X\wedge Y\in \Ww$ & Closure of $\vee, \wedge$
 \\ \hline
L2. $\; X\vee X=X$ & L2$'. \; X\wedge X=X$ & Idempotence of $\vee, \wedge$ \\ \hline
L3. $\; X\vee Y=Y\vee X$ & L3$'. \;  X\wedge Y=Y\wedge X$ & Commutativity of $\vee, \wedge$\\ \hline
L4. $\;  X\vee (Y\vee Z)=$ & L4$'. \;  X\wedge (Y\wedge Z)=$ & Associativity  of $\vee, \wedge$ \\
    \hspace*{5mm} $(X\vee Y)\vee Z$ & \hspace*{5mm} $(X\wedge Y)\wedge Z$ & \\ \hline
L5. $\; X\vee (X\wedge Y)=X$ & L5$'.\; X\wedge (X\vee Y)=X$ & Absorption between $\vee,\wedge$\\ \hline \hline
L6. $\;  X\pord Y \Longleftrightarrow$  & L6$'. \; X\dpord Y \Longleftrightarrow$ & Consistency of $\vee,\wedge$
\\ 
   \hspace*{5mm} $Y=X\vee Y$ & \hspace*{5mm} $Y=X\wedge Y$  & with partial order $\pord$
   \\ \hline
L7. $\; \flatle \vee X=X$ & L7$'. \; \flatge \wedge X=X$ & Identities of $\vee, \wedge$ \\ \hline \hline
L8. $\; \flatge \vee X=\flatge$ & L8$'. \; \flatle \wedge X=\flatle$ &
      Absorbing Nulls of $\vee, \wedge$ \\ \hline
L9. $\; X\vee (Y\wedge Z)=$  & L9$'. \; X\wedge (Y\vee Z)=$ & Distributivity of $\vee, \wedge$ \\
\hspace*{5mm} $(X\vee Y)\wedge (X\vee Z)$ & \hspace*{5mm} $(X\wedge Y)\vee (X\wedge Z)$ & \\ \hline \hline
WL10. $\; a\mgop X\in \Ww$ & WL10$'. \; a\dmgop X\in \Ww$ & Closure of $\mgop, \dmgop$
 \\ \hline
 WL11. $\; a\mgop (b\mgop X)=$ & WL11$'. \; a\dmgop (b\dmgop X)=$ & Associativity of $\mgop, \dmgop$ \\
   \hspace*{10mm} $(a\mgop b)\mgop X$ & \hspace*{10mm} $(a\dmgop b)\dmgop X$ & \\ \hline
 WL12. $\; a \mgop (X\vee Y)=$ & WL12$'. \; a \dmgop (X\wedge Y)=$ & Distributive  scalar-vector   \\
 \hspace*{10mm} $a \mgop X\vee a \mgop Y$ & \hspace*{10mm} $a \dmgop X\wedge a \dmgop Y$ & mult over vector sup/inf
 \\ \hline
 WL13. $\; (a\vee b) \mgop X=$ & WL13$'. \; (a\wedge b) \dmgop X=$ & Distributive scalar-vector  \\
 \hspace*{10mm} $a \mgop X\vee b \mgop X$ & \hspace*{10mm} $a \dmgop X\wedge b \dmgop X$ & mult over scalar sup/inf\\ \hline
 WL14. $\; \mgid \mgop X=X$ & WL14$'. \; \dmgid \dmgop X=X$ & Scalar Identities
 \\ \hline
WL15. $\; \vsetle \mgop X=\flatle$ & WL15$'. \; \vsetge \dmgop X=\flatge$ & Scalar Nulls
 \\ \hline
 WL16. $\; a \mgop \flatle=\flatle$ & WL16$'. \; a \dmgop \flatge=\flatge$ & Vector Nulls
 \\ \hline
\end{tabularx}
\label{tb-wlat}
 \end{table*}

\subsection{Vector and Signal Operators on Weighted Lattices}


We focus on CWLs whose underlying set  is a \emph{space} $\Ww=\fun(\flatdom,\vset)$
 of \emph{functions} $f:\flatdom \rightarrow \vset$
with values  from a clodum
$(\vset ,\vee,\wedge,\mgop,\dmgop)$ of scalars
as in Examples~\ref{ex-cloda}(a),(b),(c).
Such functions include $n$-dimensional vectors if $\flatdom =\{1,2,...,n\}$
or $d$-dimensional signals of continuous ($\flatdom=\REAL^d$) or discrete domain ($\flatdom=\INT^d$).
Then, we extend \emph{pointwise} the  supremum, infimum, and scalar multiplications of $\vset$ to functions:
e.g., for $F,G\in \Ww$,  $a\in \vset$ and $x\in \flatdom$, we define $(F\vee G) (x) \defineql  F(x)\vee G(x)$ and
$(a\mgop F) (x)  \defineql  a\mgop F(x)$.
Further, the scalar operations $\mgop$ and $\dmgop$, extended pointwise to functions,
distribute over any suprema and infima, respectively.
%
If the clodum $\vset$ is  self-conjugate,
then we can extend the conjugation $\glconj{(\cdot)}$ to functions $F$  pointwise: $\glconj{F}(x) \defineq \glconj{(F(x))}$.



Elementary increasing operators on
$\Ww$ are those that act as \textbf{vertical translations}
(in short V-translations)  of functions.
Specifically, pointwise `multiplications' of functions $F\in \Ww$
by scalars $a\in \vset$ yield the \emph{V-translations} $\trop _a$
and \emph{dual V-translations} $\trop' _a$,  defined by
$[\trop_a (F)](x) \defineql  a\mgop F(x)$ and
$[\trop'_a (F)](x) \defineql  a\dmgop F(x)$.
A function operator $\psi$ on $\Ww$ is called \textbf{V-translation invariant}
if it commutes with any V-translation $\trop$, i.e.,
$
\psi \trop  = \trop \psi .
$ Similarly for dual translations.

Every function $F(x)$ admits a representation as a supremum
of V-translated impulses placed at all points
or as infimum of dual V-translated impulses:
\beq
F(x) = \bigvee _{y\in \flatdom} F(y)\mgop  \dimpls_y(x)
= \bigwedge _{y\in \flatdom} F(y)\dmgop  \eimpls_y(x)
\label{sigimprep}
\eeq
where $\dimpls_{y} (x)=\mgid$ at $x=y$ and $\vsetle$ else, whereas
$\eimpls_{y} (x)=\dmgid$ at $x=y$ and $\vsetge$ else.
By using the 
V-translations and the representation of functions with impulses,
we can build more complex increasing operators.
%
%
We define operators $\dlop$ as
\textbf{dilation V-translation invariant  (DVI)} and
 operators $\erop$ as
\textbf{erosion V-translation invariant  (EVI)} iff
for any $c_i\in \vset,\; F_i \in \Ww$
\beq
\mathrm{DVI:}\; \dlop (\bigvee _{i}c_i\mgop F_i) = \bigvee _{i} c_i \mgop \dlop ( F_i),
\quad
\mathrm{EVI:}\; \erop (\bigwedge _{i}c_i\dmgop F_i) = \bigwedge _{i} c_i \dmgop \erop ( F_i)
\label{vtidilfunop}
\eeq

The structure of a DVI or EVI operator's output is
simplified if we express it via the operator's impulse
responses. Given a dilation  $\dlop$ on $\Ww$, its \textbf{impulse response map}
is the map $H:\flatdom \rightarrow \fun (\flatdom,\vset)$ defined at each $y\in \flatdom$
as the output function $H(x,y)$ from $\dlop$ when the input is the impulse $\dimpls_y(x)$.
Dually, for an erosion operator $\erop$ we  define its
 \textit{dual impulse response map} $H'$ via its outputs when excited by dual impulses:
 for $x,y\in \flatdom$
\beq
H(x,y)  \defineq  \dlop (\dimpls_y)(x),
\quad H'(x,y)  \defineq  \erop (\eimpls_y)(x) 
\label{impresp}
\eeq
Applying a DVI operator $\dlop$ or an EVI operator $\erop$
to (\ref{sigimprep}) and using
the definitions in (\ref{impresp}) yields the following unified representation,
which is proven  in \cite{BCOQ01,Mara94a} for the max-plus case and in \cite{Mara05a} for the more general max-$\mgop$ and max-$\dmgop$ cases.
\begin{Theorem} \label{th-devtirep}
(a)~An operator $\dlop$ on $\Ww$ is DVI
  iff its output can be expressed as
\beq
\dlop (F)(x) = \bigvee _{y\in \flatdom} H(x,y)\mgop F(y)
\label{tvdil}
\eeq
(b)~An operator $\erop$ on $\Ww$ is EVI
iff its output can be expressed as
\beq
\erop (F)(x) = \bigwedge _{y\in \flatdom} H'(x,y)\dmgop F(y)
\label{tvero}
\eeq
\end{Theorem}

On signal spaces, 
 the operations (\ref{tvdil}) and (\ref{tvero}) are \emph{shift-varying nonlinear convolutions}.

\paragraph{\bf Weighted Lattice of Vectors:}
\label{sc-cwlvec}

 Consider now the nonlinear vector space $\Ww =\vset ^n$, equipped
with the pointwise partial ordering $\vct{x} \leq \vct{y}$,
supremum $\vct{x}\vee \vct{y}=[x_i\vee y_i]$, and infimum
$\vct{x}\wedge \vct{y}=[x_i\wedge y_i]$ between any vectors
$\vct{x},\vct{y}\in \Ww$. Then, $(\Ww , \vee , \wedge, \mgop, \dmgop )$ is
 a complete weighted lattice.
Elementary increasing operators are the \emph{vector V-translations}
$\trop _a (\vct{x}) = a\mgop \vct{x}=[a\mgop x_i]$
and their duals $\trop' _a (\vct{x}) = a\dmgop \vct{x}$,
which `multiply' a scalar $a$ with a vector $\vct{x}$  elementwise.
A vector transformation on $\Ww$ is called (dual) V-translation invariant
if it commutes with any vector (dual) V-translation.
By defining as `impulses' 
the impulse vectors
$\vct{q}_j=[\dimpls_j(i)]$ and their duals $\vct{q}'_j=[\eimpls_j(i)]$,
where the index $j$ signifies the position of the identity,
each vector $\vct{x} = [x_1,...,x_n]^T$
has a representation as a max  of V-translated impulse
vectors or as a min of V-translated dual impulse vectors.
%
More complex examples of increasing operators on such vector spaces
are the max-$\mgop$   and the
min-$\dmgop$ `multiplications'  of a matrix $\mtr{A}$
with an input vector $\vct x$,
\beq
\dlop _{\mtr A} (\vct{x}) \defineq \mtr{A} \mxgmp \vct{x},
\; \; \;
\erop _{\mtr A} (\vct{x}) \defineq \mtr{A} \mngmp \vct{x}
\label{vecdilero}
\eeq
which are the prototypes of any vector transformation
that obeys a sup-$\mgop$ or an inf-$\dmgop$ superposition.
%
%
\begin{Theorem} \label{th-devtirepvec} \mbox{\rm \cite{Mara13,Mara17}}
(a)~Any vector transformation on the complete weighted lattice
$\Ww=\vset ^n$ is DVI
iff it can be represented as a matrix-vector max-$\mgop$ product
$\dlop _{\mtr A}(\vct x)=\mtr A\mxgmp \vct x$ where
$\mtr{A}=[a_{ij}]$ with $a_{ij}=[\dlop (\vct{q}_j)]_i$, $i,j=1,\dots,n$.
\\
(b)~Any vector transformation  on $\vset ^n$ is EVI
iff it can be represented as
a matrix-vector min-$\dmgop$ product  $\erop _{\mtr A}(\vct x)=\mtr A \mngmp \vct x$ where
$\mtr{A}=[a_{ij}]$ with $a_{ij}=[\erop (\vct{q}'_j)] _i$.
\end{Theorem}

Note that the above theorem also holds for vector transformations between CWLs of different dimensionality,
say from $\vset ^n$ to $\vset ^m$, in which case the corresponding matrix $\mtr A\in \vset ^{m\times n}$ is rectangular. Given such a vector dilation
$\dlop(\vct{x})=\mtr{A}\mxgmp \vct{x}: \vset ^n \rightarrow \vset ^m$, 
there corresponds a unique   erosion $\erop: \vset ^m \rightarrow \vset ^n$ (equal to the residual operator $\dlop ^\sharp$)
so that $(\dlop,\erop)$ is a \emph{vector adjunction}, i.e.
$\dlop (\vct x)\leq \vct y \Longleftrightarrow \vct x \leq \erop (\vct y)$.
 We can find the adjoint vector erosion by decomposing both vector operators
 based on \emph{scalar operators} $(\sbdl,\asbdl)$ that form a \emph{scalar adjunction} on $\vset$:
 \beq
\sbdl (a,v)\leq w \Longleftrightarrow v \leq \asbdl (a,w)
\label{scalaradj}
\eeq
If we use as scalar `multiplication'  a commutative binary operation
$\sbdl (a,v)=a\mgop v$ 
that is a dilation on $\vset$,  its  scalar adjoint erosion becomes
\beq
\asbdl (a,w)= \sup \{ v\in \vset : a\mgop v\leq w\}
\label{sadjerop}
\eeq
which is a (possibly non-commutative) binary operation on $\vset$.
Then,
the original vector dilation $\dlop (\vct x)=\mtr A \mxgmp \vct x$ is decomposed as
\vspace{-2mm}
\beq
[ \dlop (\vct x)] _i =\bigvee_{j=1}^n \sbdl (a_{ij},x_j)=\bigvee_{j=1}^n a_{ij}\mgop x_j, \quad i=1,...,m
\label{vdlopij}
\eeq
whereas its adjoint  vector erosion (i.e. the  residual $\dlop^\sharp$ of $\dlop$) 
is decomposed as
\beq
[ \dlop^\sharp (\vct{y})]_j=[ \erop (\vct{y})]_j=\bigwedge _{i=1}^m \asbdl (a_{ij},y_i), \quad j=1,...,n
\label{vadjeropij}
\eeq
The latter can be written as a min-$\asbdl$ matrix-vector multiplication
\beq
\erop (\vct y) = \mtr A^T \mnasbdmp \vct y
\label{vadjerop}
\eeq
where the symbol $\mnasbdmp$ denotes the following nonlinear  product
of a matrix  $\mtr{A}=[a_{ij}]$ with a matrix $\mtr{B}=[b_{ij}]$:
\[
[ \mtr{A} \mnasbdmp\mtr{B}]_{ij} \defineq \bigwedge _{k} \asbdl (a_{ik},b_{kj})
\]
Further, if $\vset=(\vee,\wedge,\mgop,\dmgop)$ is a \emph{clog}, 
then  $\asbdl (a,w)=\glconj{a} \dmgop w$ and hence
\beq
\erop (\vct{y}) = \conjtranmtr{\mtr A} \mngmp \vct y,
\quad [ \erop (\vct y)] _j=\bigwedge _{i=1}^m \glconj{a_{ij}}\dmgop y_i, \quad j=1,...,n
\label{vadjeropclog}
\eeq
where $\conjtranmtr{\mtr A}=[\glconj{a_{ji}}]$ is the \emph{adjoint} (i.e. conjugate transpose) of $\mtr A=[a_{ij}]$.
%

\paragraph{\bf Weighted Lattice of Signals:}
\label{sc-cwlsig}

Consider the set $\Ww=\fun (\flatdom, \vset )$ of all
signals $f:\flatdom \rightarrow \vset$ with domain $\flatdom=\REAL^d$ or $\INT^d$ and values from $\vset$.
The signal translations are the operators
$\trop _{y,v}(f)(x)=f(x-y)\mgop v$ and their duals.
A signal operator on $\Ww$
is called \emph{(dual) translation invariant\/} iff it commutes with any
such (dual) translation.
This translation-invariance contains both a vertical translation
and a horizontal translation (shift).
%
Consider now operators $\fdlop$ on $\Ww$ that are dilations and translation-invariant.
Then $\fdlop$ is both DVI in the sense of (\ref{vtidilfunop}) and shift-invariant. We call such operators
\textbf{dilation translation-invariant (DTI)} systems.
Applying $\fdlop$ to an input signal $f$
decomposed as supremum of translated impulses yields its output as the sup-$\mgop$ convolution $\sgcnv$
of the input with the system's impulse response $h=\fdlop (\dimpls )$,
where $\dimpls (x)=\mgid$ if $x=0$ and $\vsetle$ else:
\beq
\fdlop (f)(x)=(f\sgcnv h)(x)=\bigvee _{y\in \flatdom} f(y)\mgop h(x-y)
\eeq
Conversely, every sup-$\mgop$ convolution is a DTI system.
As done for the vector operators, we can also build  signal operator pairs $(\fdlop,\ferop)$ that form adjunctions.
Given $\fdlop$ we can find its adjoint $\ferop$  from scalar adjunctions $(\sbdl,\asbdl)$.
Thus, by (\ref{scalaradj}) and (\ref{sadjerop}), if $\sbdl (h,f)=h\mgop f$,
the adjoint  signal erosion becomes
\beq
\ferop (g)(y)=\bigwedge _{x \in \flatdom} \asbdl[h(x-y),g(x)]
\eeq
Further, if $\vset$ is a clog, then
\beq
\ferop (g)(y)=\bigwedge _{x \in \flatdom} g(x) \dmgop \glconj{h}(x-y)
\eeq
%

\subsection{CWL Generalizations of Tropical  Lines and Planes}

\begin{figure}[!h]
\centering
\subfigure[Max-plus line]
{\includegraphics[width=0.3\columnwidth]{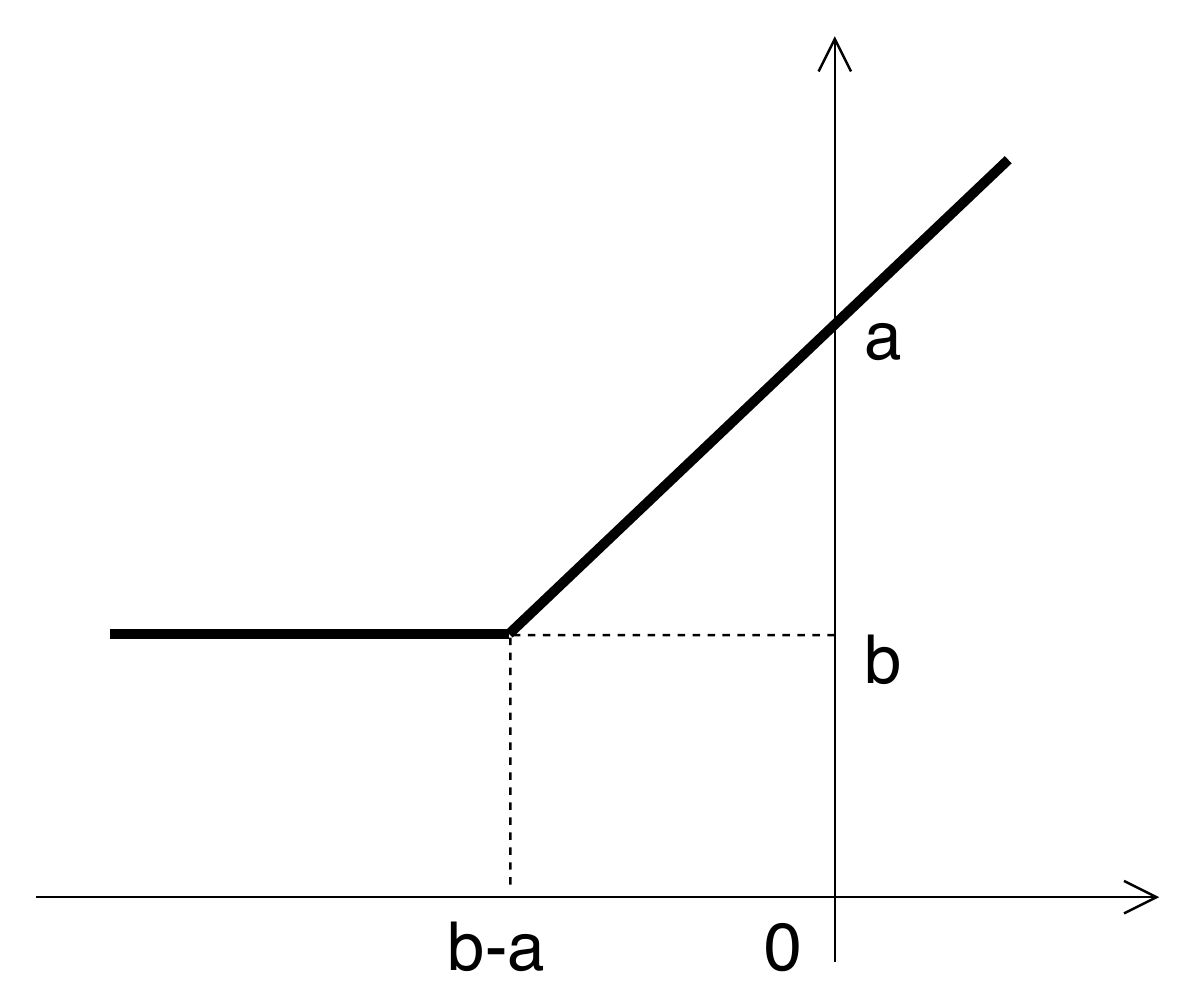}}
\subfigure[Max-product line]
{\includegraphics[width=0.3\columnwidth]{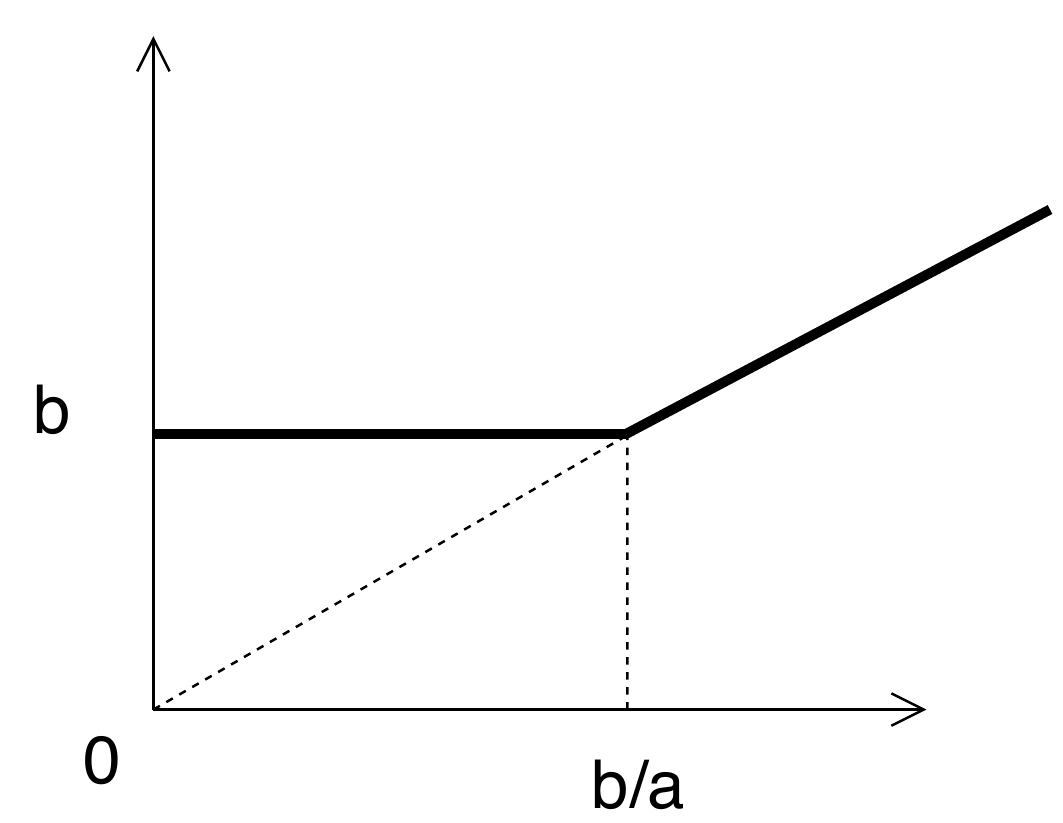}}
\subfigure[Max-min line]
{\includegraphics[width=0.3\columnwidth]{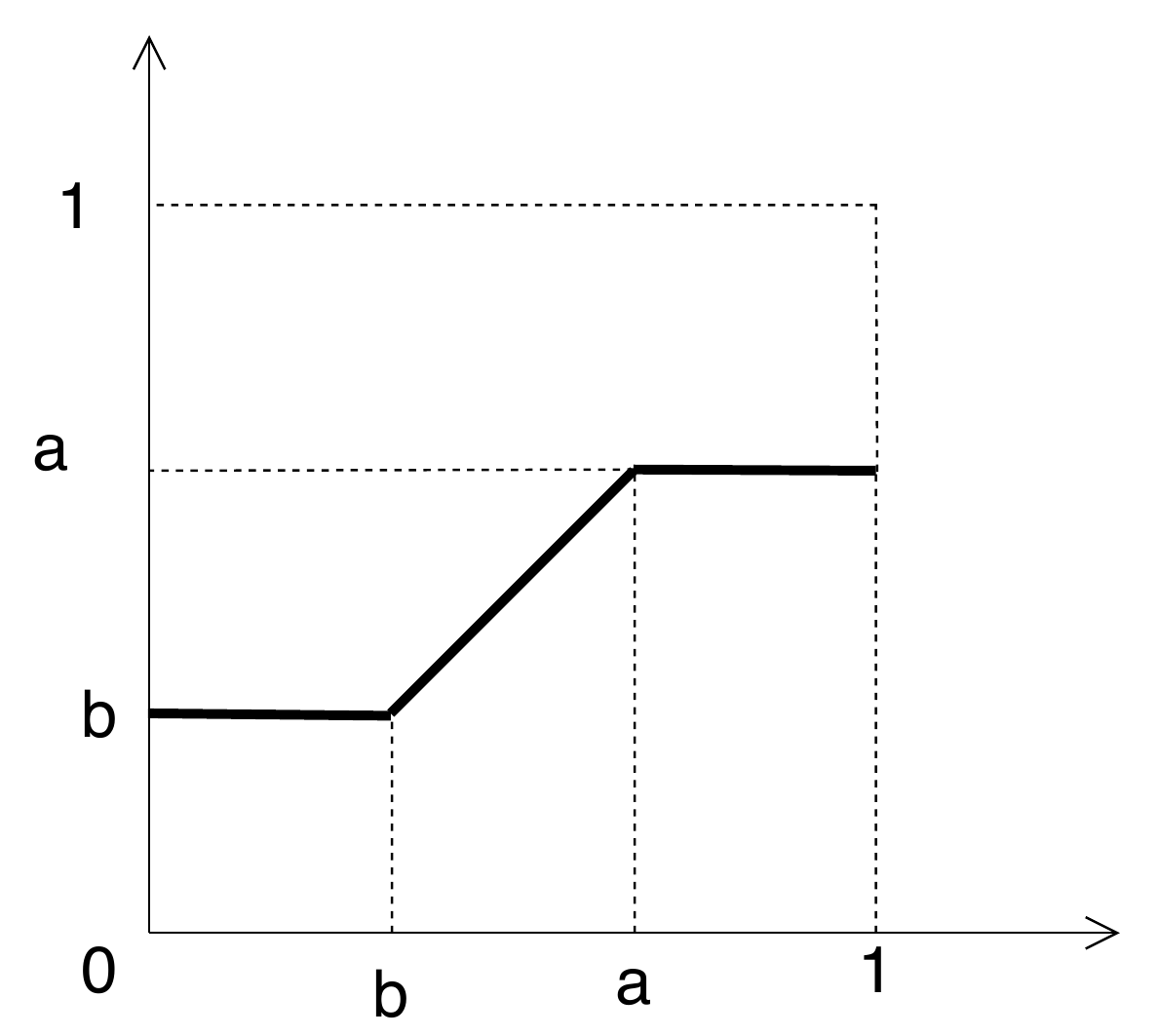}}
\caption{Max-$\mgop$ tropical lines $y=\max(a\mgop x,b)$:
(a)~$y=\max(a+x,b)$, (b)~$y=\max(a\cdot x,b)$, (b)~$y=\max(a\wedge x,b)$.
}
\label{fg-t-genline}
\end{figure}

In the same way that weighted lattices generalize max-plus morphology and extend it to other types of clodum arithmetic,
we can extend the basic objects of max-plus tropical geometry (i.e. tropical lines and planes) to other max-$\mgop$  geometric objects.
For example, over a clodum $(\Kk,\vee, \wedge, \mgop, \dmgop)$, we can generalize max-plus tropical
 lines $y=\max(a+x,b)$ as $y=\max(a\mgop x,b)$ and similarly tropical planes as $z=\max(a\mgop x,b\mgop y,c)$.
 Figure~\ref{fg-t-genline} shows some generalized tropical lines where the $\mgop$ operation is sum, product, and min.

Further, we can generalize max-plus halfspaces (\ref{max+halfsp}) to max-$\mgop$ tropical halfspaces:
\beq
\Tt (\vct a,\vct b)\defineq \{  \vct x\in \vset^n:
\vct a^T \mxgmp \left[ \begin{array}{c} \vct x \\ \mgid\end{array} \right] \leq
\vct b^T \mxgmp \left[ \begin{array}{c} \vct x \\ \mgid\end{array} \right] \}
\eeq
Examples of max-plus tropical halfspaces are shown in Fig.~\ref{fg-t-halfsp2} and Fig.~\ref{fg-polyhedron}.
The slopes of their bounding line segments or faces are either zero or equal to 1.
Max-product halfspaces can give boundaries that are piecewise-linear but have arbitrary slopes.
Max-min halfspaces have piecewise-linear boundaries with more corner points or edges;
see example in Fig.~\ref{fg-t-genline}(c).

Finally, a totally different generalization results if we replace the  `multiplication' $\mgop$ in a generalized tropical line with the
(log-sum-exp) softmin operation of Example~\ref{ex-cloda}(d), in which case the line segments of a tropical line will become smooth exponential curves.


\section{Solving Max-$\boldsymbol{\mgop}$ Equations and Optimization}
\label{sc-solmaxeqn}

\subsection{$\ell_p$  Optimal Subsolutions of Max-$\boldsymbol{\mgop}$ Equations}

Consider a scalar clodum  $(\vset,\vee,\wedge, \mgop, \dmgop)$,
a matrix  $\mtr{A}\in \vset ^{m\times n}$ and a vector $\vct{b}\in \vset ^m$.
The set of solutions of the max-$\mgop$ equation
\beq
\mtr{A} \mxgmp \vct{x} = \vct{b}
\label{mxgeq}
\eeq
over $\vset$ is either empty or forms an idempotent semigroup under vector $\vee$, because if $\vct x_1,\vct x_2$ are two solutions then
$\vct x_1\vee \vct x_2$ is also a solution.
%
A related problem in applications of max-plus algebra to scheduling is
when a vector $\vct x$  represents start times,
 a vector $\vct b$ represents finish times, and the matrix $\mtr{A}$ represents processing
 delays. Then, if $\mtr A \mxgmp \vct x  = \vct b$ does not have an exact solution,
 it is possible to find the optimum $\vct x$ such that we
 minimize a norm of the earliness subject to zero lateness.
 We generalize this problem from max-plus to max-$\mgop$ algebra. The optimum
 will be the solution of the following constrained minimization problem:
\beq
\mathrm{Minimize} \; \| \mtr A \mxgmp \vct x -\vct b \|_p \; \;  \mathrm{s.t.} \; \;
\mtr{A} \mxgmp \vct x \leq  \vct b
\label{mxgminiz}
\eeq
where the norm $||\cdot ||_p$ is any $\ell_p$  norm with $p=1,2,\dots,\infty$.
While the two above problems have been solved in \cite{Cuni79}
for the max-plus  case and for $p=1$ or $p=\infty$, 
we provide next a more general result  using adjunctions  for the
general case when $\vset$ is  just a clodum or a general clog and $||\cdot ||_p$ is any Minkowski norm.

\begin{Theorem} (\cite{Mara17}) \label{th-mxgeqminiz} \ Consider a vector dilation
$\dlop (\vct x)=\mtr A \mxgmp \vct x: \vset ^n\rightarrow \vset ^m$
over a clodum $\vset$ and let $\erop$ be its adjoint vector erosion.
(a)~If Eq.~(\ref{mxgeq}) has a solution, then
\beq
\hat{\vct x} =\erop(\vct b)  = \mtr A^T \mnasbdmp \vct b = [ \bigwedge _{i=1}^m \asbdl (a_{ij},b_i)]
\label{mxgminizsolclod}
\eeq
 is its greatest solution, where $\asbdl$ is the scalar adjoint erosion of $\mgop$ as in (\ref{sadjerop}).
 \\
(b)~If $\vset$ is a clog, the  solution (\ref{mxgminizsolclod}) becomes
\beq
\hat{\vct x}  = \conjtranmtr{\mtr A} \mngmp \vct b =[ \bigwedge _{i=1}^m \glconj{a_{ij}}\dmgop b_i]
\label{mxgminizsolclog}
\eeq
(c)~The solution to the optimization problem (\ref{mxgminiz}) for any $\ell_p$  norm $||\cdot ||_p$
is generally (\ref{mxgminizsolclod}), or  (\ref{mxgminizsolclog}) in the  case of a clog.
\end{Theorem}

A main idea  for solving (\ref{mxgminiz}) is to consider vectors $\vct x$
that are \emph{subsolutions} in the sense that $\dlop (\vct x)=A \mxgmp \vct x \leq \vct b$ and find the greatest such subsolution
$\hat{\vct x}=\erop (\vct b)$,
which  yields either the greatest exact solution of (\ref{mxgeq}) or an optimum  subsolution
in the sense of (\ref{mxgminiz}).
This creates
a lattice projection onto the max-$\mgop$ span of the columns of $\mtr A$
via the opening $\dlop (\erop (\vct b))\leq \vct b$ that best approximates $\vct b$
from below. Also, note that since $\vct y=\dlop (\erop (\vct b))=[y_i]$ is the greatest lower estimate of $\vct b=[b_i]$,
$b_i-y_i$ is nonnegative and minimum for all $i$, and hence
 the norm $\| \vct b -\vct y\|_p$ is minimum for any $p=1,2,\dots, \infty$.

 As a final note, in the max-plus case  it is also possible to search and find \emph{sparse solutions} of either
the exact equation (\ref{mxgeq}) or the approximate problem (\ref{mxgminiz}), as done in \cite{TsMa19},
where sparsity here means a large number of $-\infty$ values in the solution vector.

\subsection{Projections on Weighted Lattices}

The optimal subsolution of (\ref{mxgminiz}) can be viewed in the max-plus case as a nonlinear `projection' of $\vct b$ onto the column-space of $\mtr A$ \cite{Cuni76}. To understand this, note first that
any adjunction $(\dlop, \erop)$ automatically yields two lattice projections, an opening $\opop=\dlop \erop$ and a closing $\clop=\erop \dlop$, such that
\[
\opop^2=\opop \pord \idop \pord \clop=\clop^2
\]
where the composition of two operators is written as an operator product.
We call them `projections' because, in analogy to projection operators  on linear spaces,
they preserve the  structure of the lattice space w.r.t. the partial ordering (due to their isotonicity) and they are idempotent.

Projections on idempotent semimodules\footnote{Idempotent semimodules are like vector spaces with idempotent vector `addition' $\vee$ whose vector and scalar arithmetic are defined over idempotent semirings. 
If in our definition of a weighted lattice,
one focuses only on one vector `addition', say the vector supremum, and its corresponding scalar `multiplication',
then the weaker algebraic structure becomes an idempotent semimodule over an idempotent semiring $(\Kk, \vee, \mgop)$.
This has been studied in \cite{CGQ04,GoMi08,LMS01} where often closure under infinite suprema is assumed; in such cases, an `infimum' operation can be also indirectly defined (since a complete sup-semilattice with a least element is a complete lattice), which makes the space a complete lattice, but this indirect infimum may be different than the direct (conventional) infimum of the original lattice (if it exists).}
have been studied in  \cite{CGQ04} for the general case and in more detail for the max-plus case in \cite{AGNS11}.
Let $\Xx$ be a complete idempotent semimodule, and let $\Ss$ be a subsemimodule of $\Xx$. Then a \emph{canonical projector} on $\Ss$ is defined as the nonlinear map  \cite{CGQ04}
\beq
P_{\Ss}:\Xx \rightarrow \Xx, \quad P_{\Ss}(x)\defineq \bigvee \{  v\in \Ss: v\leq  x \}
\label{projsemimod}
\eeq
Its definition implies that $P_{\Ss}$ is a lattice opening, i.e. increasing, antiextensive, and idempotent. Further,
there is a concept of `distance' on such semimodules  which allows to use a nonlinear projection theorem for best approximations.
We shall outline these ideas only for the max-plus case, i.e. for $\Xx=\EREAL^n$ viewed as complete semimodule over the complete max-plus semiring $\Rmax \cup \{ \infty\}$.
Specifically, let us consider the  \emph{Hilbert projective metric}
\beq
 d_H(\vct x,\vct y)\defineq - [(\vct x \setminus \vct y)+(\vct y \setminus \vct x)], \quad
 \vct x \setminus \vct y\defineql \max \{ a\in \EREAL: \vct x+a \leq \vct y\}
 \label{hilbpm}
\eeq
between any vectors $\vct x, \vct y\in \EREAL^n$.
Note that this is only a semimetric and for finite-valued vectors it assumes the simpler expression
(called \emph{range semimetric} in \cite{Cuni79})
\beq
 d_H(\vct x,\vct y)= \max_i (x_i-y_i) - \min_i (x_i-y_i), \quad \vct x,\vct y\in \REAL^n
 \label{hilbpmfin}
\eeq
Then, given a subsemimodule $\Ss$ of $\EREAL^n$, it follows that for any vector $\vct x \in \EREAL^n$, $P_{\Ss}(\vct x)$ is the best approximation (but not necessarily unique) of $\vct x$ by elements of $\Ss$.
Specifically \cite{CGQ04,AGNS11},  the projection $P_{\Ss}(\vct x)$ of $\vct x$ onto $\Ss$ is that element of $\Ss$ within the shortest distance from $\vct x$ than any other element of $\Ss$; i.e.,
\beq
d_H(\vct x, P_{\Ss}(\vct x))=d_H(\vct x, \Ss)
\eeq
where the distance between a vector $\vct x$ and the subspace $\Ss$ is defined by
$d_H(\vct x, \Ss)\defineql \inf \{ d_H(\vct x,\vct v):\vct v \in \Ss\}$. Note the analogy with Euclidean spaces $\REAL^n$
where the linear projection of a point $\vct x\in \REAL^n$ to a linear subspace $\Ss$ is given by the unique point $\vct y \in \Ss$ such that $\vct x-\vct y$ is orthogonal to $\Ss$.

Now, if we consider the  optimization problem (\ref{mxgminiz}) and define the subsemimodule $\Ss$ in (\ref{projsemimod}) as the max-plus span of the columns of matrix $\mtr A$, then the canonical projection of $\vct b$ onto it equals
\beq
P_\Ss(\vct b)=\mtr A \mxsmp \hat{\vct x}  = \mtr A \mxsmp \conjtranmtr{\mtr A} \mnsmp \vct b \leq \vct b
\eeq
which is a lattice opening $\dlop (\erop (\vct b))\leq \vct b$.

\subsection{$\ell_\infty$ Optimal Solution of Max-plus Equations}

The solution (\ref{mxgminizsolclog}) is the greatest subsolution of problem (\ref{mxgminiz}).
Thus, in the max-plus case (see (\ref{maxminsummpr}) for the definitions of max-plus and min-plus matrix products), $\hat{\vct x} = \conjtranmtr{\mtr A} \mnsmp \vct b$ is the optimal solution
of
\beq
\mathrm{Minimize} \; \| \mtr A \mxsmp \vct x -\vct b \|_\infty
\label{mxsminiz}
\eeq
under the constraint $\vct x\leq \vct b$.
The proof results since $\hat{\vct x}$ is the greatest solution of $\mtr A \mxsmp \vct x \leq \vct b$, as shown by Cuninghame-Green \cite{Cuni79}. It can also be directly seen from the adjunction
\beq
\mtr A \mxsmp \vct x = \dlop_{\mtr A} (\vct x) \leq \vct b \Longleftrightarrow
\vct x \leq \erop_{\mtr A} (\vct b)=\conjtranmtr{\mtr A} \mnsmp \vct b
\eeq
The following is actually a stronger result that is not biased to be a subsolution but provides the \emph{unconstrained optimal solution} of (\ref{mxsminiz}).

\begin{Theorem} (\cite{Cuni79}) \label{th-mxseqminiz} \
If $2\mu =\| \mtr A \mxsmp \hat{\vct x} -\vct b \|_\infty=\| \mtr A \mxsmp (\conjtranmtr{\mtr A} \mnsmp \vct b) -\vct b \|_\infty$ is the $\ell_\infty$ error corresponding to the greatest subsolution of $\mtr A \mxsmp \vct x =\vct b$, then
\beq
\tilde{\vct x} = \mu + \conjtranmtr{\mtr A} \mnsmp \vct b
\eeq
is the unique optimum solution of (\ref{mxsminiz}).
\end{Theorem}

The computational complexity to find both optimal solutions $\hat{\vct x}$ and $\tilde{\vct x}$ is $O(mn)$ (additions in the max-plus case),
where $m$ is the number of data and $n$ their dimensionality.

Unfortunately, the $\ell _\infty$ optimality of $\tilde{\vct x}$ does not carry over in the case of a general clodum, as shown for the max-min clodum in \cite{CuCe95}.

\section{Optimal Fitting Tropical Polynomials to Data and Shape Approximation}
\label{sc-tropregres}

Herein we apply tropical geometry and max-$\mgop$ algebra to a fundamental regression problem of approximating the shape of curves and surfaces by fitting tropical polynomials to data, sampled from their functional form possibly in the presence of noise.

\subsection{Piecewise-Linear Function Representation and Data Fitting}

Piecewise-Linear (PWL) functions $f:\REAL^n\rightarrow \REAL$ are defined as follows: (i)~Their domain is divided into a finite number of polyhedral regions  separated by linear $(n-1)$-dimensional boundaries that are hyperplanes or subsets of hyperplanes; (ii) They are affine over each region and continuous on each boundary.
Approximations with PWL functions have proven analytically and computationally very useful in many fields of science and engineering,  including splines \cite{Boor78}, nonlinear circuits and systems modeling \cite{CDK87}, machine learning \cite{Bish06,Theo15}, convex optimization \cite{BoVa04}, geometric programming \cite{BKVH07,KVY10,MaBo09}, statistics \cite{HaDu11}, and recently tropical geometry \cite{MaSt15,Viro01}.
A conventional representation of PWL functions requires
 simplicial subdivision of their domain and interpolation of the PWL function on the subdivided domain; this is local, without a closed-formula, and requires many parameters for storage and processing.
Thus, two major problems are \emph{representation}, i.e. finding a better class of functions with analytical expressions  to represent them,  and their \emph{parameter estimation} for modeling a nonlinear system or fitting some data.
Further, while these problems are well-explored in the 1D case, they remain relatively underdeveloped for multi-dimensional data.


Chua and his collaborators \cite{KaCh78,ChDe88,KaCh92} have introduced the so-called \emph{canonical  representation} for continuous PWL functions, consisting of an affine function plus a weighted sum of absolute-value affine functions (defining linear partitions) and extensively studied its application for nonlinear circuit analysis and modeling. This has the advantages over the conventional representation that
it is global, explicit, analytic, compact (smaller number of model functions and corresponding parameters), and computationally efficient (easy to store and program).
However, it is  complete only for 1D PWL functions. In higher dimensions it needs multi-level nestings of the absolute-value functions; the depth of this nesting depends on the geometry of the partitions of the domain
and the order of intersections of the partition boundaries \cite{KaCh90,GuGo91,KaCh92,LXU94,Juli03}.

Tarela et al \cite{TaMa99}, by combining their previous work \cite{TAM90} on  representing continuous PWL functions  with lattice generalizations of Boolean polynomials of lines or hyperplanes, which extended similar work by \cite{Wilk63}, with the general $f-\phi$ model for PWL functions of \cite{LiUn94}, developed a constructive way to generate min-max (and their dual max-min) combinations of affine functions which provide a complete representation of continuous PWL functions in arbitrary dimensions.
This is called the \emph{lattice representation}.
Another work for max-min representation of PWL functions is \cite{Ovch02}.
Wang \cite{Wang04} completed the construction of a canonical representation for arbitrary continuous PWL functions in $n$-dimensions by starting from the lattice presentation of \cite{TaMa99},
which is a min-max of affine functions, producing an equivalent representation as a difference of two convex functions, each being max-affine, and then converting  each max-affine function to a canonical representation that involved $n$-level nestings of absolute-value functions.

A more recent approach is to focus on the class of \emph{convex} PWL functions represented by a maximum of affine functions (i.e. hyperplanes), that are essentially max-plus topical polynomials as in (\ref{mspolynom}), and use them for data fitting; we shall call this class \emph{max-affine} functions.
Starting from early least-squares solutions \cite{Hild54,Holl79},
some representative recent approaches to solve this \emph{convex regression} problem include \cite{HaDu11,HaDu12,HKA16,KVY10,MaBo09}. In all these approaches, there is an iteration that alternates between partitioning the data domain and locally fitting affine functions (using least-squares or some linear optimization procedure) to update the local coefficients.
For a known partition the convex PWL function is formed as the max of the local affine fits.
Then, a PWL function generates a new partition which can be used to refit the affine functions and improve the estimate. As explained in \cite{MaBo09}, this iteration can be viewed as a Gauss-Newton algorithm to solve the above nonlinear least-squares problem, similar to the $K$-means algorithm.
The order $K$ of the model can be increased until some error threshold is reached.
Interesting and promising generalizations of the above max-affine representation for convex functions include works that use softmax instead of max, via the \emph{log-sum-exp} models for convex and log-log convex data
\cite{HKA16,CGP19a,CGP19b}.
Other iterative approaches for convex PWL  data fitting include \cite{ToVi12}.
For additional references, we refer the reader to the bibliography in the above works.

Next, we focus on convex PWL regression via the max-affine model, which has a tropical interpretation,
and propose
a direct \emph{non-iterative} and \emph{low-complexity} approach to estimate its parameters
by using the optimal solutions of max-plus (or max-$\mgop$) equations of Sec.~\ref{sc-solmaxeqn}.
We note that the max-affine representation is not limited to PWL functions only, because we can represent any convex function as a supremum of a (possibly infinite) number of affine functions via
 the Fenchel-Legendre transform \cite{Fenc49,Rock70,Luce10}.
Closely related ideas are based on morphological slope transforms that offer generalizations of this result, either as lattice-theoretic adjunctions that can also yield approximate representations  \cite{HeMa97,Mara94a,Mara95} or as multi-valued Legendre transforms in case of differentiable non-convex or non-concave functions \cite{DoBo94}.
%

\subsection{Optimal Fitting Tropical Lines and Planes}

We first examine a classic problem in machine learning,
fitting a line to data by minimizing an error norm, in the light of tropical geometry.
Given data $(x_i,f_i)\in \REAL^2$, $i=1,...,m$, if we wish to fit a Euclidean line $y=ax+b$
by minimizing the $\ell_2$ error norm $\| \vct f-a\vct x-b\|_2$ where $\vct f=[f_i]$ and $\vct x=[x_i]$,
 the optimal  solution (\emph{least squares estimate - LSE}) for the parameters $a,b$ is
\beq
\hat{a}_\textrm{LS}=\frac{m\sum_ix_if_i-(\sum_ix_i)(\sum_if_i)}{m\sum_i(x_i)^2-(\sum_ix_i)^2}, \quad \hat{b}_\textrm{LS}=\frac{1}{m}\sum_i (f_i-\hat{a}_\textrm{LS}x_i)
\eeq
Suppose now we wish to fit a general tropical line $p(x)=\max(a\mgop x,b)$ by minimizing some $\ell_p$ error norm. The equations to solve for finding the optimal parameter vector $\vct w=[a,b]^T$ become:
\beq
\underbrace{\left[ \begin{array}{cc} x_1 & \mgid \\
\vdots & \vdots \\ x_m & \mgid \end{array} \right] }_{\mtr X} \mxgmp
\underbrace{\left[ \begin{array}{c} a \\ b \end{array} \right]}_{\vct w} =
\underbrace{\left[ \begin{array}{c} f_1 \\ \vdots \\ f_m \end{array}\right]}_{\vct f}
\label{mxgeqn-tropline}
\eeq
By Theorem~\ref{th-mxgeqminiz},  the optimal (min $\ell_p$ error) subsolution  for any clodum arithmetic is
\beq
\hat{\vct w}=\left[ \begin{array}{c} \hat{a} \\ \hat{b} \end{array} \right]
= \mtr X^T \mnasbdmp \vct f
= \left[ \begin{array}{c} \bigwedge_{i} \asbdl (x_i,f_i)
      \\ \bigwedge_{i}\asbdl (\mgid,f_i) \end{array} \right]
\label{mxgsolgen-tropline}
\eeq
  where $\asbdl$ is the scalar adjoint erosion (\ref{sadjerop}) of $\mgop$.
  This vector $\hat{\vct w}$  yields (after min-$\asbdl$ `multiplication' with $\mtr X^T$) the \emph{greatest lower estimate (GLE)} of the data $\vct f$.
 If $\vset$ is a clog, like in the max-plus and max-product case, then
   $\asbdl (x_i,f_i)=\glconj{x_i} \dmgop f_i$.
 Next we write in detail the solution for the tropical line for the three special cases where
   the scalar arithmetic is based either on the max-plus clog\footnote{To cover all cases of combining finite and infinite scalar numbers in the max-plus clog $(\EREAL, \vee,\wedge,+,+')$, we should write the subtractions $f_i-x_i$ in (\ref{mxgsolclod3-tropline}) as $f_i+'(-x_i)$.}, or the max-product clog, or the max-min clodum
   (the shapes of the corresponding lines are shown in Fig.\ref{fg-t-genline}):
\beq
(\hat{a}, \; \hat{b})  = \left\{ \begin{array}{ll}
 \bigwedge_i f_i-x_i, \; \bigwedge_i f_i), & \mbox{\rm max-plus}\; (\mgop = +) \\
 \bigwedge_i f_i/x_i, \; \bigwedge_i f_i), & \mbox{\rm max-times}\; (\mgop = \times) \\
  \bigwedge_i \max([f_i\geq x_i],f_i), \; \bigwedge_i f_i), & \mbox{\rm max-min}\; (\mgop = \wedge)
  \end{array} \right.
\label{mxgsolclod3-tropline}
\eeq
where $[\cdot ]$ denotes Iverson's bracket in the max-min case.
Thus, the above approach allows to optimally fit (w.r.t. any $\ell_p$ error norm) general tropical lines to arbitrary data from below.
In addition, for the \emph{max-plus} case we can obtain the best (unconstrained) approximation with a tropical line that yields the smallest $\ell_\infty$ error. This \emph{minimum max absolute error (MMAE)} solution
is, by Theorem~\ref{th-mxseqminiz},
\beq
\tilde{\vct w} = \hat{\vct w}+\mu, \quad \mu  =\frac{1}{2}\| \mtr X \mxsmp \hat{\vct w} -\vct f \|_\infty
=\frac{1}{2}\| \mtr X \mxsmp (\conjtranmtr{\mtr X} \mnsmp \vct f) -\vct f \|_\infty
\label{mmaesol-tropline}
\eeq

\begin{figure}[!h]
\centering
\subfigure[T-line with Gaussian Noise]
{\includegraphics[width=0.45\columnwidth]{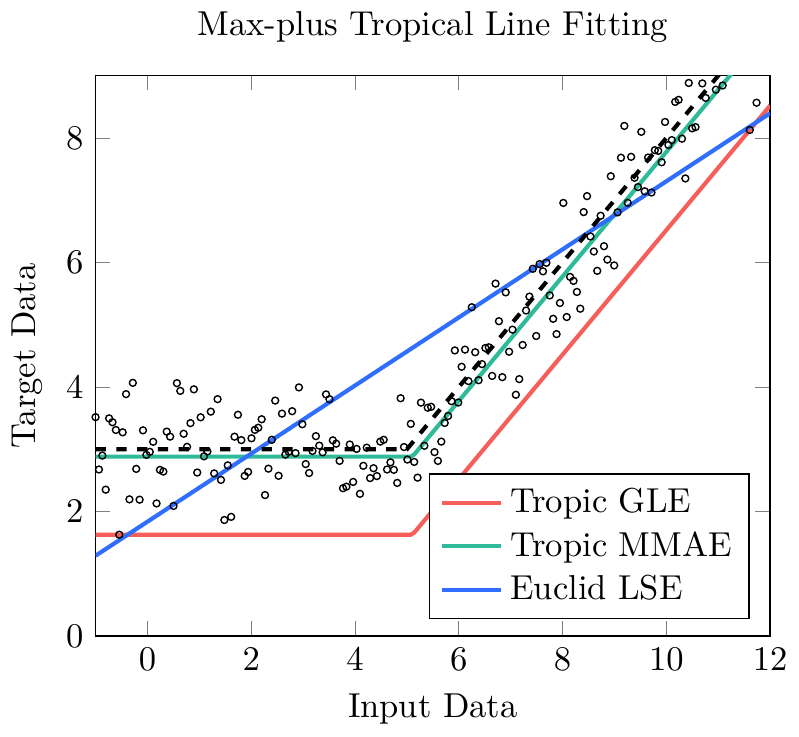}}
\hspace{5mm}
\subfigure[T-line with Uniform Noise]
{\includegraphics[width=0.45\columnwidth]{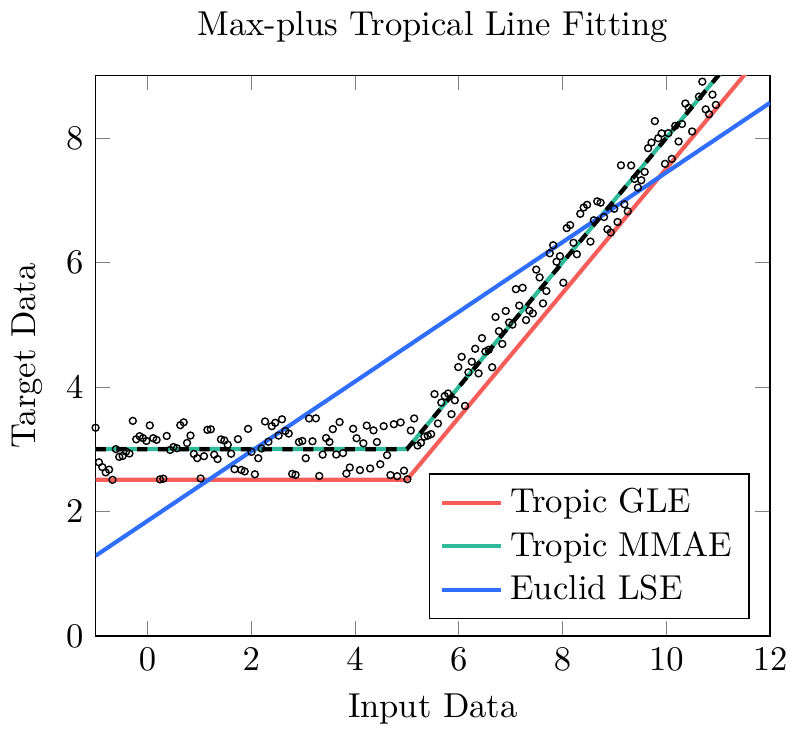}}
\caption{(a)~Optimal fitting via (\ref{mxgsolclod3-tropline}) or (\ref{mmaesol-tropline}) of a max-plus tropical line $y=\max(x-2,3)$ (shown in black dashed curve) to data from the line corrupted by additive i.i.d. Gaussian noise $\sim {\mathcal N}(0,0.25)$. Blue line: Euclidean line fitting via least squares.
Red line: best subsolution (GLE). Green line: best unconstrained (MMAE) solution.
(b)~Same experiment as in (a) but with uniform noise $\sim \text{Unif}[-0.5,0.5]$. Best viewed in color.}
\label{fg-tropic-line-fit}
\end{figure}

\begin{Example} {\rm
Suppose we have $m=200$ data observations $(x_i,f_i)$ from the tropical line
	$y = \max(x - 2, 3)$,
where the 200 abscissae $x_i$  were uniformly spaced in $[-1, 12]$ and their corresponding values $f_i=y_i+\epsilon_i$ are contaminated with two different types of zero-mean noise i.i.d. random variables $\epsilon_i$, Gaussian noise $\sim {\mathcal N}(0,0.25)$ and uniform noise $\sim \text{Unif}[-0.5,0.5]$.
Figure~\ref{fg-tropic-line-fit} shows the two optimal solutions (\ref{mxgsolclod3-tropline}) and  (\ref{mmaesol-tropline}) for fitting a max-plus tropical line, superimposed with the least-squares Euclidean line fit. The parameter estimates and errors are in Table~\ref{tb-linefit}.
\begin{table}[!h]
\centering
\begin{tabular}{|l|c|c|c|c|} \hline
Line fit Method & $\| \mathrm{error}\|_{\text{RMS}}$ & $\| \mathrm{error}\|_\infty$ & $\hat{a}$ & $\hat{b}$ \\ \hline \hline
Tropical GLE & 0.598 & 0.988 & -2.492 & 2.509 \\ \hline
Tropical MMAE & 0.288 & 0.494 & -1.998 & 3.003 \\ \hline
Euclidean LSE & 0.968 & 2.135  & 0.560 & 1.849 \\ \hline
\end{tabular}
\caption{Errors and parameter estimates for optimal fitting of a max-plus tropical line $y=\max(x-2,3)$ both via a least-squares Euclidean line fit and via the tropical constrained (GLE) and unconstrained (MMAE) solutions, to data corrupted by uniform noise.}
\label{tb-linefit}
\end{table}
}
\end{Example}

The above approach and tropical solution can also be extended to fitting planes.
Specifically, we wish to fit a general max-$\mgop$ tropical plane $p(x,y)$
\beq
p(x,y) = \max(a\mgop x,b\mgop y,c)
\eeq
to given data $(x_i,y_i,f_i)\in \REAL^3$, $i=1,...,m$,
where $f_i=p(x_i,y_i)+\textrm{error}$,
by minimizing some $\ell_p$ error norm.
As in (\ref{mxgeqn-tropline}),
the equations to solve for finding the optimal parameters $\vct w=[a,b,c]^T$ become:
\beq
\underbrace{\left[ \begin{array}{ccc} x_1 & y_1 &\mgid \\
\vdots & \vdots & \vdots \\ x_m & y_m & \mgid \end{array} \right] }_{\mtr X} \mxgmp
\underbrace{\left[ \begin{array}{c} a \\ b \\ c \end{array} \right]}_{\vct w} =
\underbrace{\left[ \begin{array}{c} f_1 \\ \vdots \\ f_m \end{array}\right]}_{\vct f}
\label{mxgeqn-tropplane}
\eeq
If we can accept subsolutions, which yield approximations of the given data from below, then by Theorem~\ref{th-mxgeqminiz}  the optimal  subsolution  for any clodum arithmetic is
\beq
\hat{\vct w}=\left[ \begin{array}{c} \hat{a} \\ \hat{b} \\ \hat{c} \end{array} \right]
= \mtr X^T \mnasbdmp \vct f
= \left[ \begin{array}{c} \bigwedge_{i} \asbdl (x_i,f_i) \\
      \bigwedge_{i} \asbdl (y_i,f_i) \\
      \bigwedge_{i}\asbdl (\mgid,f_i) \end{array} \right]
\label{mxgsol-tropline}
\eeq
In the special case of \emph{max-plus} arithmetic, then $\asbdl (x_i,f_i)=f_i-x_i$ and the best subsolution (for min $\ell_p$ error) becomes
\beq
\left[ \begin{array}{c} \hat{a} \\ \hat{b} \\   \hat{c} \end{array} \right]=
\hat{\vct w} = \conjtranmtr{\mtr X} \mnsmp \vct f
=\left[ \begin{array}{cccc} -x_1 & -x_2 & \cdots & - x_m \\
-y_1 & -y_2 & \cdots & -y_m \\
0 & 0 & \cdots & 0 \end{array} \right] \mnsmp
\left[ \begin{array}{c} f_1  \\ f_2  \\ \vdots \\ f_m \end{array} \right]
=
\left[ \begin{array}{c} \bigwedge_{i=1}^m f_i-x_i  \\ \bigwedge_{i=1}^m f_i-y_i  \\
 \bigwedge_{i=1}^m f_i \end{array} \right]
 \label{mxssol-tropplane}
\eeq
Furthermore, the minimum max absolute error (MMAE) solution is given by (\ref{mmaesol-tropline}),
but the data matrix $\mtr X$ and vector $\vct f$ refer now to the plane case.

\subsection{Shape Regression by Optimal Fitting Tropical Max-plus Polynomial Curves and Surfaces}

For the \emph{max-plus} case, the above approach and solution can also be generalized to polynomial curves of higher degree and to multi-dimensional data.
We wish to fit a max-plus tropical polynomial
\beq
p(\vct x) = \max(\vct a_1^T\vct x+b_1,\vct a_2^T\vct x+b_2,\dots, \vct a_K^T\vct x+b_K)=\bigvee_{k=1}^K \vct a_k^T\vct x+b_k, \quad \vct x\in \REAL^n
\label{fit_magenpol}
\eeq
to given data $(\vct x_i,f_i)\in \REAL^{n+1}$, $i=1,...,m$, where $f_i=p(\vct x_i)+\textrm{error}$,
 by minimizing some $\ell_p$ error norm. The exact equations are
 \beq
\underbrace{ \left[ \begin{array}{cccc} \vct a_1^T\vct x_1 & \vct a_2^T\vct x_1 & \cdots & \vct a_K^T\vct x_1 \\
 \vct a_1^T\vct x_2 & \vct a_2^T\vct x_2 & \cdots & \vct a_K^T\vct x_2 \\
 \vdots & \vdots & \vdots & \vdots \\
 \vct a_1^T\vct x_m & \vct a_2^T\vct x_m & \cdots & \vct a_K^T\vct x_m
\end{array} \right] }_{\mtr X} \mxsmp
\underbrace{ \left[ \begin{array}{c}  b_1 \\ b_2 \\ \vdots \\ b_K \end{array} \right]}_{\vct w} =
\underbrace{\left[ \begin{array}{c} f_1  \\ f_2  \\ \vdots \\ f_m \end{array} \right] }_{\vct f}
 \label{fit_mageneqn}
\eeq
We assume that the slope vectors $\vct a_k$ are given and we optimize for the parameters $\{ b_k\}$.
By Theorem~\ref{th-mxgeqminiz},  the
optimal subsolution for minimum $\ell_p$ error is
\beq
\left[ \begin{array}{c}  \hat{b}_1 \\  \vdots \\ \hat{b}_K \end{array} \right]=
\hat{\vct w} = \conjtranmtr{\mtr X} \mnsmp \vct f
=\left[ \begin{array}{cccc} 
-\vct a_1^T\vct x_1 & -\vct a_1^T\vct x_2 & \cdots & -\vct a_1^T\vct x_m \\
\vdots & \vdots & \vdots & \vdots \\
-\vct a_K^T\vct x_1 & -\vct a_K^T\vct x_2 & \cdots & -\vct a_K^T\vct x_m \end{array} \right] \mnsmp
\left[ \begin{array}{c} f_1  \\ f_2  \\ \vdots \\ f_m \end{array} \right]
=
\left[ \begin{array}{c}
\bigwedge_{i=1}^m f_i-\vct a_1^T\vct x_i  \\ \vdots \\
 \bigwedge_{i=1}^m f_i-\vct a_K^T\vct x_i \end{array} \right]
\label{fit_magensolgle}
\eeq
Note that $\mtr X \mxsmp \hat{\vct w}\leq \vct f$.
Further, by Theorem~\ref{th-mxseqminiz},
the unconstrained solution that yields the minimum $\ell_\infty$ error is
\beq
\tilde{\vct w}=\mu + \hat{\vct w}, \quad \mu  =\frac{1}{2}\| \mtr X \mxsmp \hat{\vct w} -\vct f \|_\infty
\label{fit_magensolmae}
\eeq
Our assumption for known  slope vectors $\vct a_k$ does not pose a significant constraint in many cases where the
degree of the polynomial is relatively small,
in which case we assume that $\vct a_k$ are integer multiples of a slope step
or simply that they assume all integer values up to the maximum degree.
If this is not the case, another approach  is to compute the derivatives (or gradients) of the given data,
 estimate the histogram of the derivative values, and use this for automatic selection of the slope parameters.
 Or simply to cluster the data gradients using $K$-means and use the centroids of the $K$ clusters as our given slope vectors.
In both approaches, setting $b_k=-\infty$ for some $k$, removes the corresponding line or hyperplane from the max-affine combination.
Next we apply the above approaches for optimally solving two cases (1D and 2D) with examples.

\subsubsection{Optimal Fitting 1D Tropical Max-plus Polynomial Curves}
We wish to fit a max-plus tropical polynomial curve
\beq
p(x) = \max(b_{-r}-rx, \dots, b_{-1}-x,b_0,b_1+x,\dots, b_r+rx)  =\bigvee_{k=-r}^r b_k+kx
\label{fit_ma1intpol}
\eeq
of relatively low degree $r$ with $K=2r+1$ terms to given data $(x_i,f_i)\in \REAL^2$, $i=1,...,m$, where $f_i=p(x_i)+\textrm{error}$,
 by minimizing some $\ell_p$ error norm.
The tropical polynomial $p(x)$ is a maximum of straight lines with integer slopes
$a_k=k\in \REAL$ and intercepts $b_k\in \Rmax$; a null intercept means that the corresponding line does not contribute to $p(x)$.
The above PWL model is efficient if the function to approximate has both positive and  nonnegative slopes
and over the approximation interval its slopes do not exceed a relatively low polynomial degree $r$.
If it has only nonnegative slopes, then we do not include the terms with negative slopes.

Based on  the PWL model (\ref{fit_ma1intpol}), the equations to solve  for finding the optimal parameters $\{ b_k\}$ become:
\beq
\underbrace{ \left[ \begin{array}{ccccccc} -rx_1 & (1-r)x_1 & \cdots & 0 & \cdots & (r-1)x_1 & rx_1 \\
 -rx_2 & (1-r)x_2 & \cdots & 0 & \cdots & (r-1)x_2 & rx_2 \\
 \vdots & \vdots & \vdots & \vdots & \vdots & \vdots & \vdots \\
 -rx_m & (1-r)x_m & \cdots & 0 & \cdots & (r-1)x_m & rx_m
\end{array} \right] }_{\mtr X} \mxsmp
\underbrace{ \left[ \begin{array}{c}  b_{-r} \\ b_{1-r} \\ \vdots \\ b_r \end{array} \right]}_{\vct w} =
\underbrace{\left[ \begin{array}{c} f_1  \\ f_2  \\ \vdots \\ f_m \end{array} \right] }_{\vct f}
\eeq
\begin{Example} {\rm
See Fig.~\ref{fg-tropfit_circle} for a numerical example where the data to fit resulted from sampling the bottom half of the circular curve $x^2 + (y-10)^2 = 7^2$ at $m=5$ points with abscissae $(x_1,\dots,x_5)=(-5.5, -2,1.5, 4,6.5)$.
 The optimal fit was done using a tropical polynomial as in (\ref{fit_ma1intpol}) with $r=3$, i.e. a max of 7 lines,
yielding a MMAE of $\| [p(x_i)]-[y_i]\|_\infty=0.12$.
%

\begin{figure}[!h]
\centering
{\includegraphics[width=0.60\textwidth]{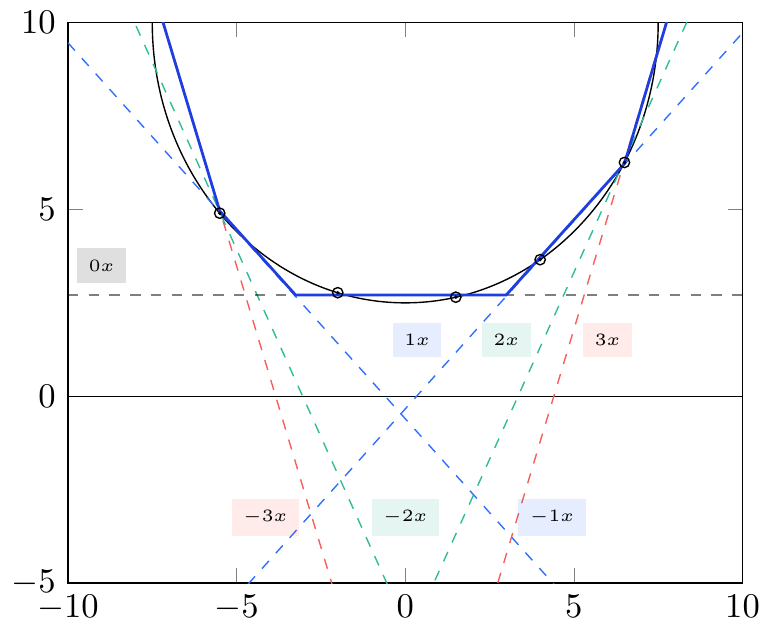}}
\caption{Piecewise-linear curve approximation of a half circle (black solid line) by interpolating 5 samples with a tropical max-plus polynomial (blue solid line). The individual lines of the PWL function are shown with dashed lines.}
\label{fg-tropfit_circle}
\end{figure}

} \end{Example}

%
%
\begin{Example} {\rm
As another 1D example, consider clean data $(x_i,y_i)$ that are $m=100$ points
with abscissae $x_i$ uniformly sampled within the interval $[-2,2]$
and ordinates $y_i=f_i=f(x_i)$ where $f(x)$ is  the convex function \cite{HKA16}
\beq
	f(x) = \max (-6x-6, \frac{x}{2}, \frac{x^5}{5} + \frac{x}{2})
\label{hoburg}
\eeq
The tropical model we are fitting is of the form
	$p(x) = \max (a_1x + b_1, ..., a_K x + b_K)$
where the slopes $a_k$ are computed using the Jenks natural breaks optimization (which is essentially a 1D $K$-means) algorithm
applied to the numerical derivatives of the data
and the intercepts $b_k$ are computed using the 1D ($n=1$) version of the
tropical fitting algorithms (\ref{fit_magensolgle}) and  (\ref{fit_magensolmae}).
See Fig.~\ref{fg-hoburg} for the curve approximations
and Table~\ref{tb-hoburg} for the corresponding errors.
\begin{figure}[!h]
\centering
\subfigure[$K=3$ lines]
{\includegraphics[width=0.45\columnwidth]{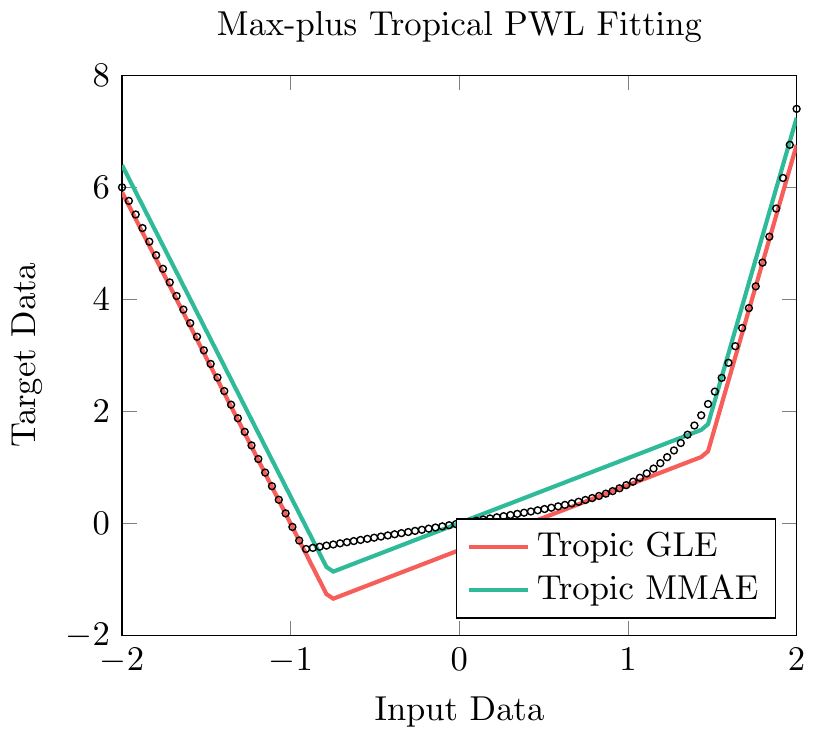}}
\hspace{5mm}
\subfigure[$K=4$ lines]
{\includegraphics[width=0.45\columnwidth]{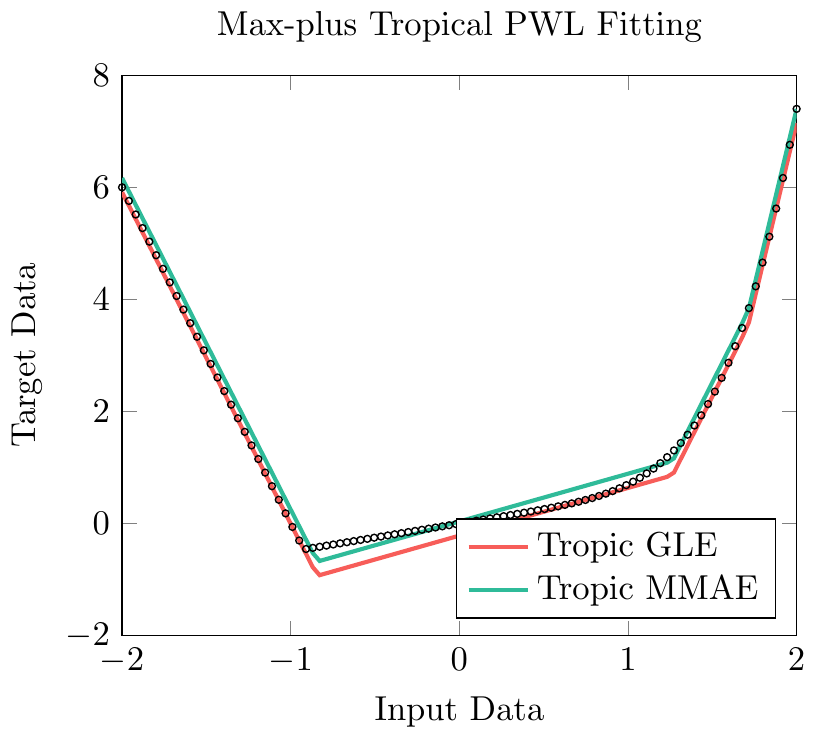}}
\centering
\subfigure[$K=5$ lines]
{\includegraphics[width=0.45\columnwidth]{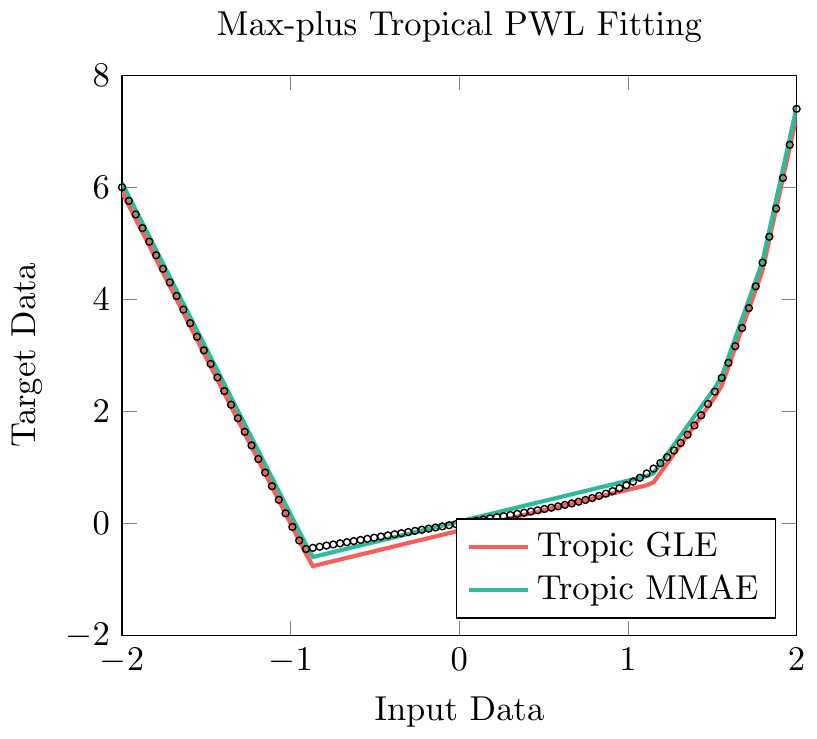}}
\hspace{5mm}
\subfigure[$K=6$ lines]
{\includegraphics[width=0.45\columnwidth]{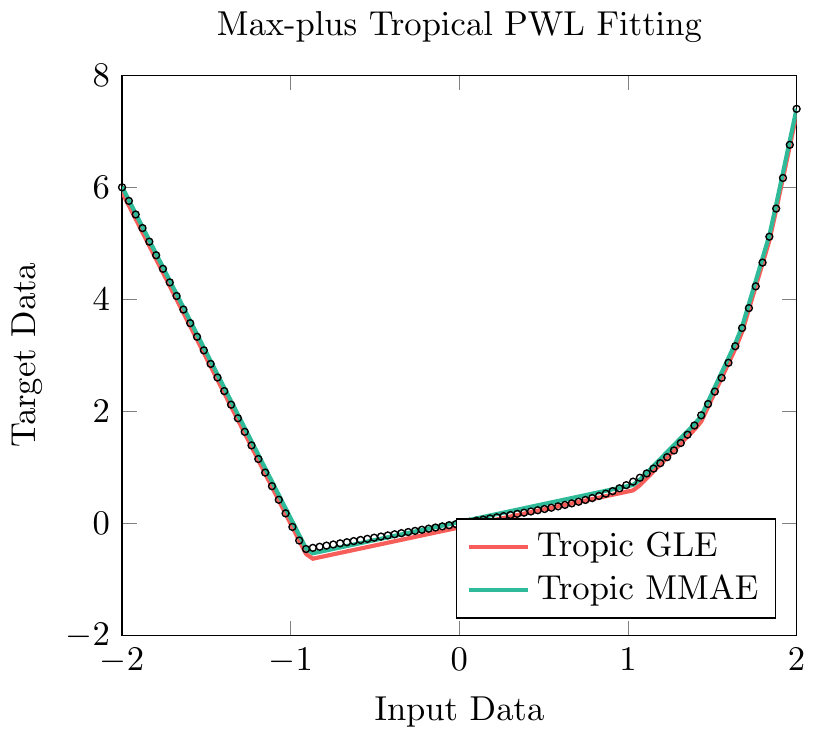}}
\caption{Fitting max-plus tropical polynomials to the function (\ref{hoburg}).
Best viewed in color.
}
\label{fg-hoburg}
\end{figure}

\begin{table}[!h]
\centering
	\begin{center}
		\begin{tabular}{c|c|c||c|c|}
			\cline{2-5}
			& \multicolumn{2}{|c||}{GLE} & \multicolumn{2}{|c|}{MMAE}\\
			\hline
			\multicolumn{1}{|c|}{$K$} & $\text{error}_{\text{RMS}}$ & $\lVert\text{error}\rVert_\infty$ & $\text{error}_{\text{RMS}}$ & $\lVert\text{error}\rVert_\infty$\\
			\hline
			\multicolumn{1}{|c|}{$3$} & $0.4101$ & $0.9671$ & $0.3535$ & $0.4836$\\
			\multicolumn{1}{|c|}{$4$} & $0.2048$ & $0.5072$ & $0.1799$ & $0.2536$\\
			\multicolumn{1}{|c|}{$5$} & $0.1230$ & $0.7226$ & $0.3004$ & $0.3613$\\
			\multicolumn{1}{|c|}{$6$} & $0.0801$ & $0.1932$ & $0.0625$ & $0.0966$\\
			\hline
		\end{tabular}
	\end{center}
	\caption{Minimum RMS error and maximum absolute error for the optimal constrained (GLE) and unconstrained (MMAE) tropical fitting of the function (\ref{hoburg}).}
\label{tb-hoburg}
\end{table}
For the case $K=6$ the estimates for the slopes and the MMAE solution for intercepts yielded
\beq
\begin{array}{rcl}
(a_1, a_2, a_3, a_4, a_5, a_6) & = & (-5.92, 0.64, 3.07, 6.43, 10.08, 14.11) \\
(b_1, b_2, b_3, b_4, b_5, b_6) & = & (-5.82, 0.03, -2.5, -7.3, -13.4, -20.83)
\end{array}
\label{hoburg_estim}
\eeq
For the case $K=3$, although our method yields about double the RMS error of the method in \cite{HKA16}, the latter
is  computationally more complex as explained later.
} \end{Example}

\subsubsection{Optimal Fitting 2D Tropical Max-plus Polynomial Surfaces}

\begin{figure}[!h]
\begin{center}
\subfigure[2D conic, GLE]
{\includegraphics[width=0.45\textwidth]{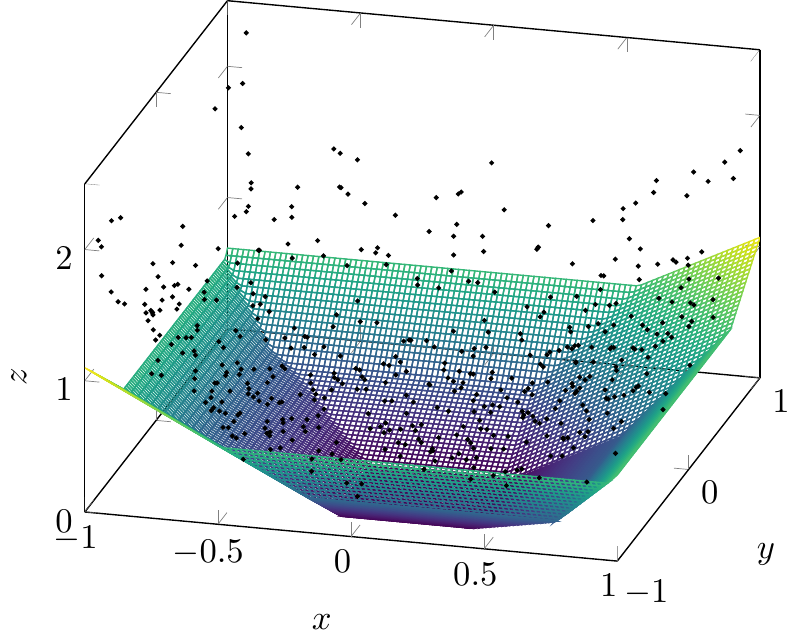}}
\subfigure[2D conic, MMAE]
{\includegraphics[width=0.45\textwidth]{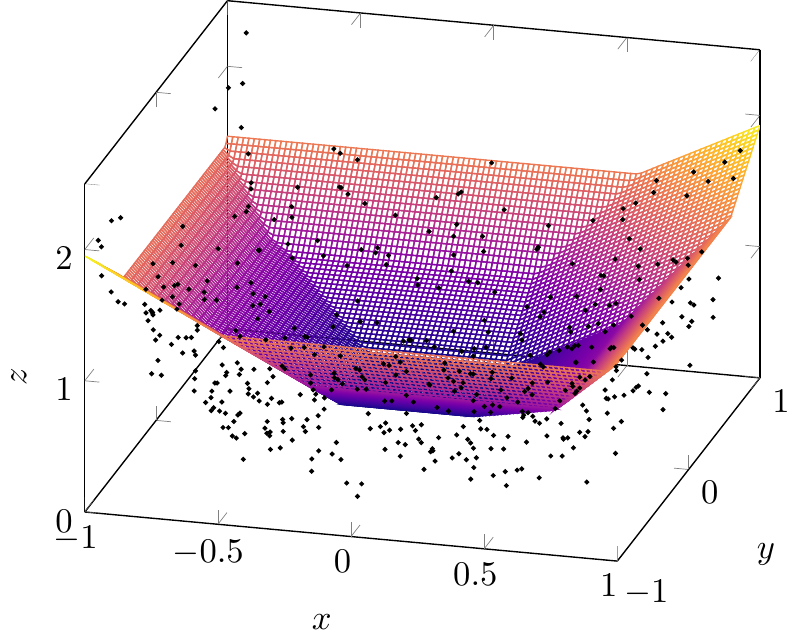}}
\end{center}
\caption{Piecewise-linear surface approximation of a noisy paraboloid with a 2D tropical max-plus conic polynomial.}
\label{fg-tropfit-paraboloid-conic}
\end{figure}

As a 2D example with known slopes, let us fit the graph surface of a symmetric max-plus tropical conic polynomial
\beq
p(x,y)=\bigvee_{0\leq |k+\ell| \leq 2, \; k\ell\geq 0}b_{k\ell}+kx+\ell y
\eeq
to given data $(x_i,y_i,f_i)\in \REAL^3$, $i=1,...,m$, where $f_i=p(x_i,y_i)+\mathrm{error}$ by minimizing some $\ell_p$ error norm.
The equations to solve for finding the optimal parameters $b_{k\ell}$, adjusted as in (\ref{fit_ma1intpol}) so that they include both negative and positive slopes, become:
\beq
\underbrace{ \left[ \begin{array}{ccccccccccc}
-2y_1 & -2x_1 & -x_1-y_1 & -y_1 & -x_1 & 0 & x_1 & y_1 & x_1+y_1 & 2x_1 & 2y_1   \\
-2y_2 & -2x_2 & -x_2-y_2 & -y_2 & -x_2 & 0 & x_2 & y_2 & x_2+y_2 & 2x_2 & 2y_2 \\
\vdots & \vdots & \vdots & \vdots & \vdots & \vdots \\
-2y_m & -2x_m & -x_m-y_m & -y_m & -x_m & 0 & x_m & y_m & x_m+y_m & 2x_m & 2y_m
\end{array} \right] }_{\mtr X} \mxsmp
\underbrace{ \left[ \begin{array}{c} b_{0,-2}\\ b_{-2,0}\\ b_{-1,-1}\\ b_{0,-1}\\ b_{-1,0}\\ b_{0,0}\\
b_{1,0}\\ b_{0,1}\\b_{1,1}\\b_{2,0}\\b_{0,2}
\end{array} \right]}_{\vct w} =
\underbrace{\left[ \begin{array}{c} f_1\\ f_2 \\ \vdots \\ f_m \end{array} \right] }_{\vct f}
\eeq
By Theorems~\ref{th-mxgeqminiz} and \ref{th-mxseqminiz}, the optimal subsolution (GLE) $\hat{\vct w}$ for minimum $\ell_p$ error and the optimal unconstrained solution $\tilde{\vct w}$ (for MMAE) equal
\beq
\hat{\vct w} = \conjtranmtr{\mtr X} \mnsmp \vct f, \quad \tilde{\vct w}=\mu + \hat{\vct w}
\eeq
where  $\mu$ is half the $\ell_\infty$ error incurred by $\hat{\vct w}$.
The GLE and MMAE solutions for the model are shown in Fig.~\ref{fg-tropfit-paraboloid-conic} for fitting data from a noisy paraboloid surface.

\begin{Example} {\rm
The data $(x_i,y_i,f_i)$ in Fig.~\ref{fg-tropfit-paraboloid-conic} are 500 observations  \cite{HaDu11}
from the noisy paraboloid surface $z = x^2 + y^2$ corrupted by a zero-mean random noise $\epsilon \sim \mathcal{N}(0, 0.25^2)$.
Thus, $f_i=f(x_i,y_i)$ where
\beq
	f(x,y)=z+\epsilon = x^2 + y^2 + \epsilon
\label{hannah}
\eeq
and the planar locations $x_i, y_i$ of the data points were drawn as i.i.d. random variables $\sim \mathrm{Unif}[-1, 1]$.
Now,
the general model we are fitting has degree $K$ and is
\beq
	p(x,y) = \max(a_1 x + b_1 y + c_1, ..., a_K x + b_K y + c_K),
\eeq
where the slopes $(a_k, b_k)$ are computed using $K$-means on the numerical gradients of the 2D data, and
the intercepts $c_k$ are computed using the tropical fitting algorithm. See Fig.~\ref{fg-hannah} for the resulting approximations
and Table~\ref{tb-hannah} for the error norms.

\begin{figure}[!h]
\centering
\subfigure[$K$=10 GLE]
{\includegraphics[width=0.45\columnwidth]{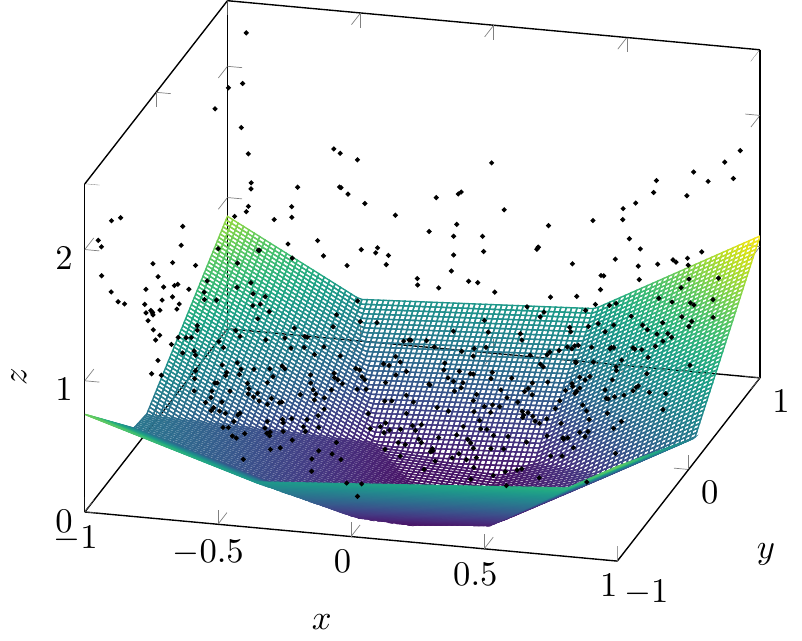}}
\hspace{5mm}
\subfigure[$K$=10 MMAE]
{\includegraphics[width=0.45\columnwidth]{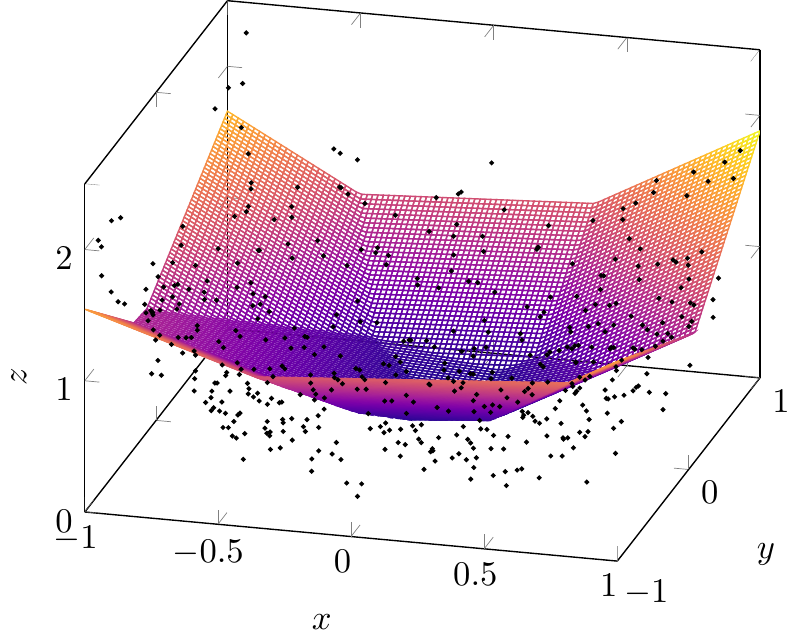}}
\centering
\subfigure[$K$=25 GLE]
{\includegraphics[width=0.45\columnwidth]{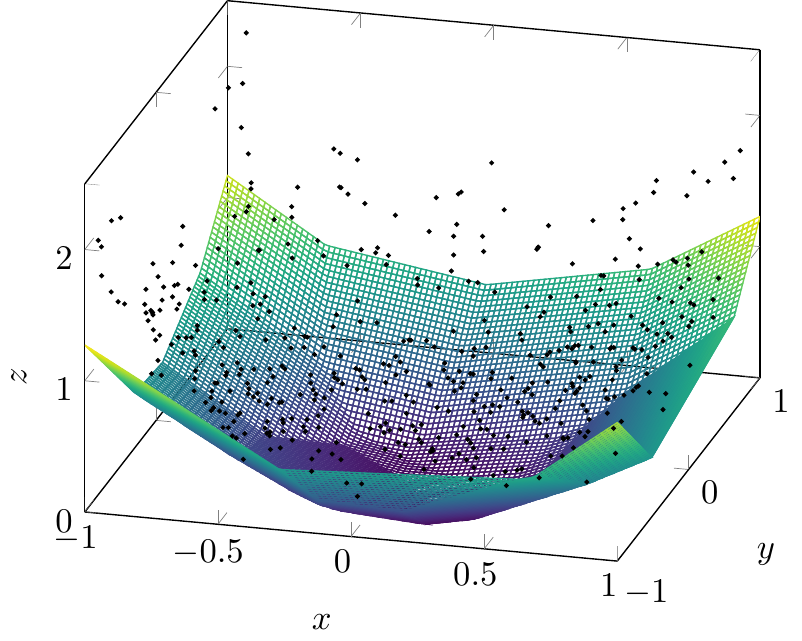}}
\hspace{5mm}
\subfigure[$K$=25 MMAE]
{\includegraphics[width=0.45\columnwidth]{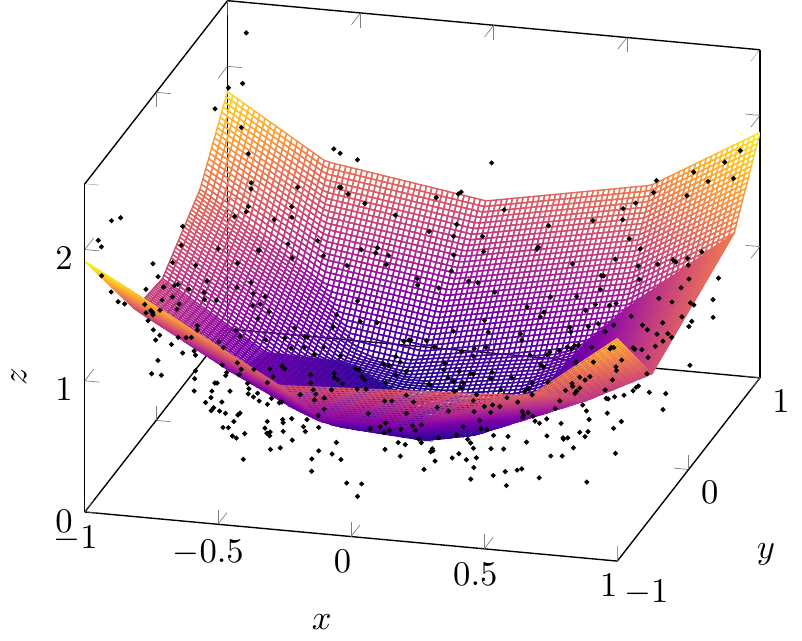}}
\caption{2D Tropical fitting using the optimal constrained (GLE) and unconstrained (MMAE) approach to data from (\ref{hannah}).
Best viewed in color.}
\label{fg-hannah}
\end{figure}

\begin{table}[!h]
\centering
%
%
\begin{tabular}{c|c|c||c|c|}
			\cline{2-5}
			& \multicolumn{2}{|c||}{GLE} & \multicolumn{2}{|c|}{MMAE}\\
			\hline
			\multicolumn{1}{|c|}{$K$} & $\text{error}_{\text{RMS}}$ & $\lVert\text{error}\rVert_\infty$ & $\text{error}_{\text{RMS}}$ & $\lVert\text{error}\rVert_\infty$\\
			\hline
			\multicolumn{1}{|c|}{$11$ (conic)} & $0.6307$ & $1.7049$ & $0.4167$ & $0.8524$\\
			\hline
			\multicolumn{1}{|c|}{$10$} & $0.6659$ & $1.6022$ & $0.3641$ & $0.8011$\\
			\multicolumn{1}{|c|}{$25$} & $0.5674$ & $1.2779$ & $0.3016$ & $0.6389$\\
			\multicolumn{1}{|c|}{$50$} & $0.5489$ & $1.3068$ & $0.3159$ & $0.6534$\\
			\multicolumn{1}{|c|}{$75$} & $0.5433$ & $1.2950$ & $0.3150$ & $0.6475$\\
			\multicolumn{1}{|c|}{$100$} & $0.5364$ & $1.2828$ & $0.3135$ & $0.6414$\\
			\multicolumn{1}{|c|}{$250$} & $0.5273$ & $1.2786$ & $0.3172$ & $0.6393$\\
			\hline
		\end{tabular}
\caption{Minimum RMS error and maximum absolute error for the optimal constrained (GLE) and unconstrained (MMAE)
 tropical fitting of the function (\ref{hannah}) using either a 2D tropical conic or $K$-term optimal fit
 whose gradients are found via $K$-means.}
\label{tb-hannah}
\end{table}
} \end{Example}

\noindent \textbf{Computational Complexity}: \\
The prevailing trend in recent methods  to fitting $m$ data points in $\REAL^{n+1}$ using as model  $n$-dimensional max-affine functions (i.e. max of $K$ hyperplanes $\vct a_k^T+b_k$),
which we view as max-plus tropical polynomials with real slope and intercept parameters, is a variety of iterative nonlinear least-squares algorithms.
 The number of model parameters is $K(n+1)$.
The traditional least-squares estimator (LSE) \cite{Hild54,Holl79} solves a quadratic program with constraints and
has a total complexity of $O((n+1)^3m^3)$. Clearly, this becomes practically intractable for large number of data points and, also, as the dimensionality increases.
In \cite{MaBo09,HKA16} the nonlinear least-squares problems is solved
 iteratively  where each iteration involves some partitioning of the data into $K$ clusters and least-squares fitting of hyperplanes over the different $K$ clusters. This has a complexity of $O((n+1)^2mi_C)$ where $i_C$ is the number of iterations until convergence; however, this least-squares partition algorithm does not always converge, and even in cases of convergence the fit to the data may be poor. To overcome this obstacle, the authors in \cite{MaBo09,HKA16}  propose running several instances of their algorithm, with different random  initializations, in order to achieve a better fit to the data.
The convex adaptive partitioning (CAP) algorithm proposed in \cite{HaDu11} has a complexity
 at $O(n(n+1)^2m \log(m) \log(\log(m)))$,
 where its most demanding part is linear regression since each least-squares fit has complexity $O((n+1)^2m)$.
 Although the CAP algorithm seems to have a slightly larger complexity than that of \cite{MaBo09}, it provides a \emph{consistent} estimator.

 In contrast, the complexity of our algorithm is dominated only by the $K$-means computation, which has a complexity of $O(Kmni_K)$, where $i_K$ is the number of $K$-means iterations. After the $K$ centroids $\vct a_k$ have been computed, our algorithm simply does a single pass over the  data for the tropical regression to find the $b_k$, with total complexity $O(Kmn)$. Therefore, the overall complexity of our tropical regression algorithm (both via the GLE and the MMAE criteria) is $O(Kmni_K)$. In general, assuming that the data have some clustering structure, the required number of $K$-means iterations to find the slopes  is small and thus our algorithm can be  considered `linear' in practice. As such, in non-pathological cases, we can assume that the product $Ki_K$ is significantly smaller than $m$ and can be treated as a constant, resulting in an overall complexity of $O(mn)$, thus improving on the CAP algorithm bound \cite{HaDu11}, and greatly improving on the traditional LSE. In terms of performance, as long as the number of clusters is not too small (and thus there are many elements in the cluster that are not adequately represented by the centroid), then the tropical algorithm will produce good PWL fits to the data (as is evident for the 1D case in Fig.~\ref{fg-hoburg} for $K = 6$ and the 2D case in Fig.~\ref{fg-hannah} for $K=25$.)

\section{Conclusions}

(Max-plus) Tropical Geometry and (Weighted) Mathematical Morphology share a common idempotent semiring arithmetic, which also has a dual counterpart.
Both can be extended and generalized using max-$\mgop$ algebra over Complete Weighted Lattices (CWLs) which are nonlinear vector spaces.
The CWL framework allows for optimal solutions (using adjunctions and lattice projections) for general max-$\mgop$ systems of equations,
which are applied to optimal fitting of tropical lines or hyperplanes to data.
This tropical regression provides convex piecewise-linear (PWL) approximations to curves and surfaces with max-affine functions at a linear complexity with respect to the number of data and their dimension,
which is significantly lower than the complexity of least-square estimates for PWL shape regression.

\begin{acknowledgement}
 The authors wish to thank V. Charisopoulos  for insightful discussions on tropical geometry and neural networks.
\end{acknowledgement}

\bibliographystyle{plain}
\bibliography{bibliography_dswl_maxplus_tropgeo}

\end{document}